\documentclass[10pt,journal]{IEEEtran}
\usepackage{amsmath,amsfonts}
\usepackage{algorithmicx}
\usepackage[ruled, linesnumbered]{algorithm2e}
\usepackage{array}
\usepackage{textcomp}
\usepackage{stfloats}
\usepackage{url}
\usepackage{verbatim}
\usepackage{graphicx}
\usepackage{cite}
\usepackage{subfigure}
\usepackage{marvosym}
\usepackage{boldline}
\usepackage{soul}
\usepackage{multirow}
\usepackage[hidelinks]{hyperref}
\usepackage{diagbox}
\usepackage{amssymb}
\usepackage{paralist,bm}
\usepackage{makecell}
\usepackage{xcolor}
\usepackage{booktabs}
\usepackage{times}
\usepackage{epsfig}
\usepackage{enumitem}
\usepackage{ragged2e}
\usepackage{algpseudocode}
\graphicspath{{figs/}}

\def\ie{\emph{i.e.~}}

\newenvironment{packed_itemize}{
	\begin{itemize}
		\setlength{\itemsep}{1pt}
		\setlength{\parskip}{0pt}
		\setlength{\parsep}{0pt}
	}{\end{itemize}}

\begin{document}

\title{Boosting Cross-Domain Point Classification via Distilling Relational Priors from 2D Transformers}

\author{Longkun Zou${^*}$, Wanru Zhu${^*}$,
Ke Chen\textsuperscript{\Letter},~\IEEEmembership{Member,~IEEE}, 
Lihua Guo\textsuperscript{\Letter},~\IEEEmembership{Member,~IEEE},
Kailing Guo,~\IEEEmembership{Member,~IEEE},
Kui Jia,~\IEEEmembership{Member,~IEEE}
and
Yaowei Wang,~\IEEEmembership{Member,~IEEE}

\IEEEcompsocitemizethanks{
\IEEEcompsocthanksitem This work is supported in part by the Guangdong Pearl River Talent Program (Introduction of Young Talent) under Grant No. 2019QN01X246, the Guangdong Basic and Applied Basic Research Foundation under Grant No. 2023A1515011104 and the Major Key Project of Peng Cheng Laboratory under Grant No. PCL2023A08. \textit{(Longkun Zou and Wanru Zhu contributed equally to this work.) (Corresponding author: Ke Chen; Lihua Guo.)}
\IEEEcompsocthanksitem L. Zou, W. Zhu, L. Guo and K. Guo are with the School of Electronic and Information Engineering, South China University of Technology, Guangzhou, 510641, China.
\IEEEcompsocthanksitem L. Zou is an intern at the Peng Cheng Laboratory, Shenzhen 518000, China.
\IEEEcompsocthanksitem K. Chen and Y. Wang are with the Peng Cheng Laboratory, Shenzhen 518000, China.
\IEEEcompsocthanksitem K. Jia is with the Chinese University of Hong Kong, Shenzhen (CUHK-Shenzhen), Shenzhen 518000, China.}}

\markboth{Journal of \LaTeX\ Class Files,~Vol.~14, No.~8, August~2021}%
{Shell \MakeLowercase{\textit{et al.}}: A Sample Article Using IEEEtran.cls for IEEE Journals}

\IEEEpubid{0000--0000/00\$00.00~\copyright~2021 IEEE}

\maketitle

\begin{abstract}

Semantic pattern of an object point cloud is determined by its topological configuration of local geometries.
Learning discriminative representations can be challenging due to large shape variations of point sets in local regions and incomplete surface in a global perspective, which can be made even more severe in the context of unsupervised domain adaptation (UDA).
In specific, traditional 3D networks mainly focus on local geometric details and ignore the topological structure between local geometries, which greatly limits their cross-domain generalization.
Recently, the transformer-based models have achieved impressive performance gain in a range of image-based tasks, benefiting from its strong generalization capability and scalability stemming from capturing long range correlation across local patches.
Inspired by such successes of visual transformers, we propose a novel \textbf{R}elational \textbf{P}riors \textbf{D}istillation (RPD) method to extract relational priors from the well-trained transformers on massive images, which can significantly empower cross-domain representations with consistent topological priors of objects.
To this end, we establish a parameter-frozen pre-trained transformer module shared between 2D teacher and 3D student models, complemented by an online knowledge distillation strategy for semantically regularizing the 3D student model. 
Furthermore, we introduce a novel self-supervised task centered on reconstructing masked point cloud patches using corresponding masked multi-view image features, thereby empowering the model with incorporating 3D geometric information.
Experiments on the PointDA-10 and the Sim-to-Real datasets verify that the proposed method consistently achieves the state-of-the-art performance of UDA for point cloud classification. The source code of this work is available at {\color{blue}\url{https://github.com/zou-longkun/RPD.git}}.


\end{abstract}


\section{Introduction}
\IEEEPARstart{T}{he}  point cloud is one of the popular 3D shape representations, with broad applications in robotics, drones, autonomous driving, \textit{etc}.
Semantic pattern of an object point cloud is determined by its topological configuration of local geometries.
Recent advances in point cloud semantic analysis \cite{PointNet, PointNet++, DGCNN, PointCNN, PointConv, PointMLP, KPConv} have been largely driven by synthetic point clouds generated from CAD models (such as those in the ModelNet \cite{Modelnet} and the ShapeNet \cite{Shapenet}), which typically have noise-free point-based surface in local regions and a complete topological structure.
Real-world point cloud data generated from RGB-D scanned by real-time depth sensors (such as the ScanNet \cite{ScanNet} and the ScanObjectNN \cite{ScanObjectNN}) typically contains noises and occlusion, making it to suffer from large shape variations of point sets in local regions and incomplete surface in a global perspective.
Such geometric variations can cause performance degradation when testing the network on a domain different from the training ones. More often, labels in the test domain may be unavailable due to high annotation costs, which is the situation we are interested in and can be formulated as the problem of unsupervised domain adaptation (UDA). 

\begin{figure}[t]
  \centering
   \includegraphics[width=0.95\linewidth]{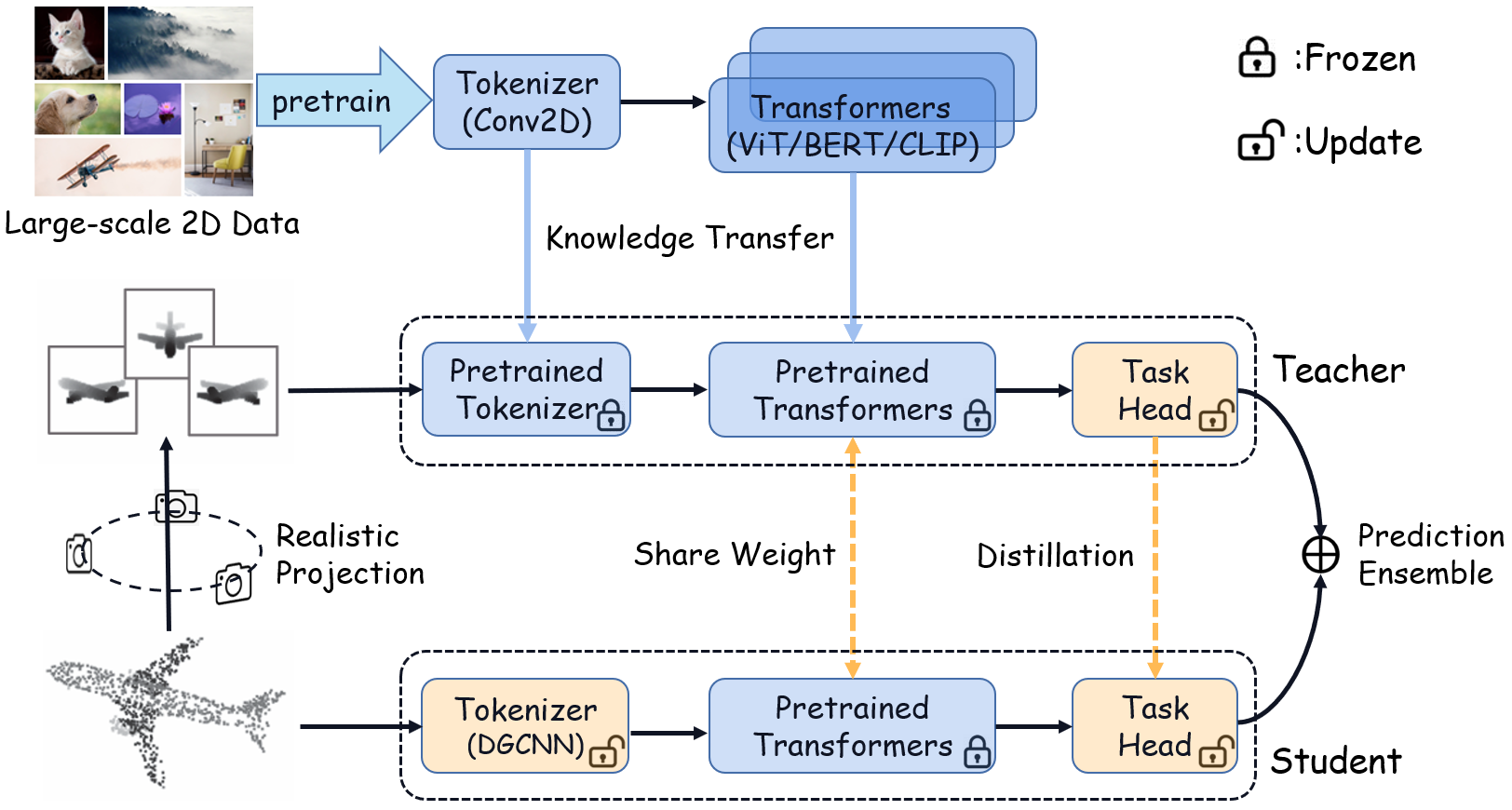}
   \caption{Illustration of the proposed relational prior distillation framework (RPD) method. We leverage the relational priors of one pretrianed 2D Transformer model to boost the 3D Transfermer encoder via sharing a parameter-frozen pretrained Transformer module and employing an online knowledge distillation strategy as semantic regularization for 3D student model. An ensemble of the knowledge from the two modalities can effectively improve the generalization of point cloud representations to close domain gap.}
   \label{fig:first}
\end{figure}

\IEEEpubidadjcol 
Unsupervised domain adaptation on point clouds is recently attracted increasing attention in \cite{PointDAN, DefRec, GAST, GAI, GLRV, QS3} started since the pioneering PointDAN \cite{PointDAN}.
In general, these point-based UDA methods can be mainly categorized into 
 two group of algorithms to bridge domain gap: domain adversarial training based \cite{PointDAN} and self-supervised learning based \cite{GAST, DefRec, GLRV, GAI}.
The former employs domain adversarial training to explicitly enforce indistinguishable features between point clouds from different domains using domain discriminators. Its main ideas are borrowed from the image-based UDA \cite{DANN, ADDA, MDD, MCD, NRMT}, which can be unstable and has a potential risk of damaging the intrinsic structures of target data discrimination in feature space, resulting in a suboptimal adaptation.

The latter mechanism achieves implicit domain alignment by incorporating self-supervised regularization pretext tasks aimed at capturing domain-invariant geometric patterns alongside semantic representation learning.
The underlying motivation is that well-designed self-supervised tasks shared across domains can facilitate the learning of features with similar properties, which typically have a certain degree of cross-domain invariance.
A diverse set of well-designed designed self-supervised tasks are proposed, such as rotation angle classification and deformation location \cite{GAST}, deformation reconstruction \cite{DefRec}, scaling-up-down prediction and 3D-2D-3D projection reconstruction \cite{GLRV}, and global implicit fields learning \cite{GAI}. 
The PDG \cite{PDG} utilized the DGCNN \cite{DGCNN} or the PointNet \cite{PointNet} to encode part-level features, which are used as a dictionary to describe other features from local parts with a linear weighting strategy.
However, existing point-based UDA algorithms mainly often prioritize feature alignment while overlooking the topological structure between local geometries, which greatly limits their cross-domain generalization capabilities.

Recently, transformer-based models have demonstrated remarkable success across various image-based tasks, following the ``pretrain-and-finetune" paradigm, which can be attributed to their robust generalization capability and scalability, stemming from their ability to capture long-range correlations across local patches. 
Nonetheless, achieving proficiency in discerning topological relationships among local parts necessitates pre-training on extensive datasets.
Mainstream point cloud networks, constrained by limited training data, leading to usage of shallow architectures to evade over-fitting, but this compromises their scalability and hampers their capacity to capture robust generalization features. 
Consequently, these networks struggle to effectively implement the ``pretrain-and-finetune" paradigm and typically require training from scratch.
While certain approaches, such as the PCT \cite {PCT} and the Point Transformer \cite{PointTransformer}, integrate the typical Transformer architecture into the 3D domain to deepen networks and enhance scalability, their efficacy remains contingent upon access to substantial labeled 3D data.
In contrast, acquiring and annotating 2D data is comparatively straightforward, with vast datasets readily available online, numbering in the millions or even billions (e.g., the ImageNet \cite{imagenet}, the COCO \cite{coco}, the CLIP \cite{CLIP}). 
Leveraging these extensive 2D datasets, 2D transformer based networks exhibit superior aptitude in capturing topological relationships among local parts.
This prompts a pivotal question: \textit{Can we harness the abundant relational priors ingrained in pre-trained 2D Transformer-based models to bolster the generalization capabilities of 3D models and mitigate domain shift?} 
Affirmative answers to this question would not only bridge the 2D and 3D modalities but also diminish the heavy reliance on expensive collection and annotation of 3D data for model pre-training.

To harness the rich relational priors ingrained in pre-trained 2D Transformer-based models, we propose a simple yet effective knowledge distillation scheme with the standard teacher-student distillation workflow, whose concept is depicted in Fig. \ref{fig:first}.
Initially, both the teacher and student models share the frozen parameters of the standard Transformer module where the parameters of most block layers are fixed and only the last few block layers are fine-tuned.
Moreover, we adopt an online knowledge distillation strategy, alternating between training the teacher and student models throughout the training process. We employ the KL-divergence loss function to align the predicted logits of the teacher and student models, enhancing cross-modal knowledge transfer and serving as semantic regularization for the 3D student model.
Additionally, recognizing that sole reliance on 2D knowledge might inadequately capture 3D geometric information, we introduce a self-supervised task of reconstructing masked point clouds from projected multi-view images. In this way, the model's ability to capture geometric information is enhanced. During inference, we ensemble predictions from both modalities.
Our method achieves state-of-the-art performance on two public benchmark datasets (\ie PointDA-10 \cite{PointDAN} and Simt-to-Real \cite{MetaSets}), which validates the effectiveness of our proposed method. In summary, our approach innovatively bridges the gap between 2D and 3D domains by leveraging the strength of Transformer-based attention mechanisms, which excel in modeling the relationships between local parts. This not only improves the robustness and generalization of 3D networks but also provides a practical solution to the data scarcity challenge in the 3D domain.
Our main contributions in this study are as follows: 
\begin{packed_itemize}
\item This paper proposes a novel scheme for unsupervised domain adaptation on object point cloud classification, which bridges domain gap via distilling relational priors from well-learned 2D transformers into 3D domains to enhance 3D feature representation.
\item Technically, we propose a simple but effective cross-modal knowledge transfer method, in which a parameter-frozen pretrained transformer module is shared between the 2D teacher and 3D student model and an online knowledge distillation strategy is adopted as a semantic regularization for 3D student model.
\item Meanwhile, we design a novel self-supervision task that reconstructs masked point cloud patches with corresponding masked multi-view image features to enhance the model’s ability to capture geometric information.
\item Experiments on two public UDA benchmarks verify that the proposed method consistently achieves the best performance of UDA for point cloud classification.
\end{packed_itemize}

\section{Related Works}
\begin{figure*}
  \centering
  \includegraphics[width=0.9\linewidth]{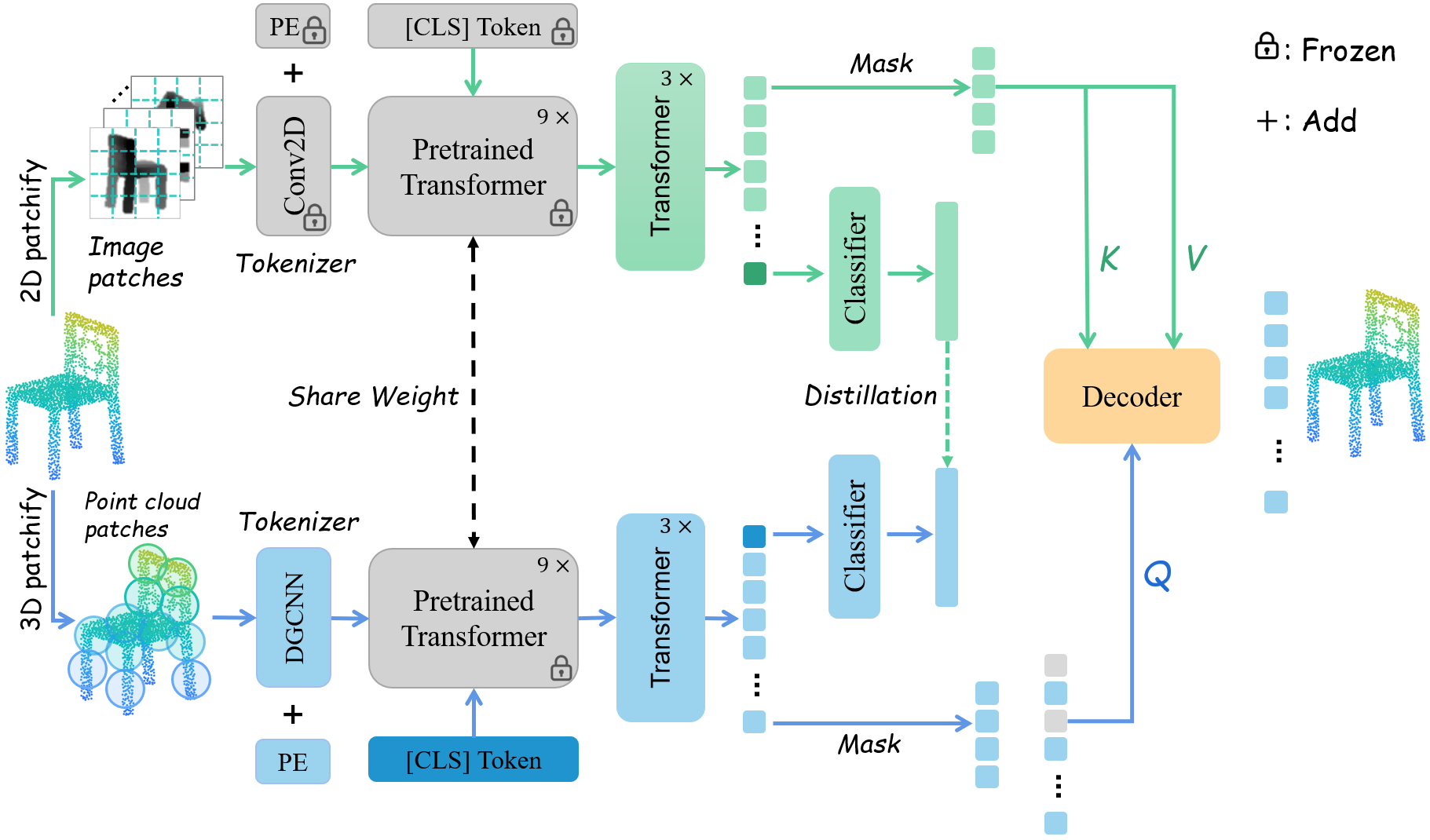}
  \caption{
  Overview of our proposed relational priors distillation framework, which adheres to a standard teacher-student distillation workflow. Both the 2D teacher model and the 3D student model include Patchify, Tokenizer, and several Transformer encoder layers. For 2D teacher model, we project the point cloud into 10 single-channel depth maps via the Realistic Projection Pipeline introduced by PointClip v2 \cite{PointCLIPv2}, and then ``patchify" these depth maps into $10 \times 14 \times 14$ image patches as input to the 2D Tokenizer (\ie Conv2D). Tokens from the 2D Tokenizer and a [CLS] token are fed into the Transformer encoder. For 3D student model, we ``patchify" the point cloud into 27 groups via Farthest Point Sampling (FPS) as input to the 3D Tokenizer (\ie DGCNN \cite{DGCNN}). Tokens from the 3D Tokenizer and a [CLS] token are fed into the Transformer encoder. The two modalities are processed independently by a siamese Transformer encoder parametrized by a MAE \cite{MAE} pre-trained ViT \cite{vit}. During training, we randomly mask a pairs of point cloud token features and image token features with a huge fraction of 0.85. The decoder consists of a sequence of multi-head cross-attention (MCA) and multi-heat self-attention (MSA) layers and predicts missing patches in the point cloud with unmasked image token features. PE means the position encoding. Gray boxes indicate parameters are frozen, while blue, green and orange boxes indicate parameters can be updated. (Best viewed in color).
  }
  \label{fig:model}\vspace{-0.3cm}
\end{figure*}

\subsection{Deep Networks for Point Clouds}
In recent years, deep neural network architectures for point clouds have been extensively studied. Existing methods can be roughly divided into three major categories: view-based \cite{3DSegPCN, MVCNN, GVCNN, MVC, MVTN, SimpleView, MvNet} and voxel-based \cite{Minkowski-CNN, SSCNs, VoxNet}, and point-based point cloud processing methods \cite{DGCNN, PointMLP, PointNet, PointNet++}.

View-based methods project the point cloud into images of multiple views and process them with various variants of 2D CNNs. The pioneering work MVCNN \cite{MVCNN} consumes the multi-view images rendered from multiple virtual camera poses and obtains global shape features through cross-view max-pooling. GVCNN \cite{GVCNN} proposes a three-level hierarchical correlation modeling framework, which adaptively groups multi-view feature embeddings into separate clusters. RotationNet \cite{RotationNet} treats viewpoint indices as learnable latent variables and tends to jointly estimate object poses and semantic categories. MVTN \cite{MVTN} introduces differentiable rendering techniques to implement adaptive regression of optimal camera poses in an end-to-end trainable manner. SimpleView \cite{SimpleView} naively project raw points onto image planes and set their pixel values according to the vertical distance. MvNet \cite{MvNet} proposes a multi-view vision-prompt to bridge the gap between 3D data and 2D pretrained models. 
Although view-based methods have shown dominant performance in various shape recognition tasks [2], [25], [26], acquiring views requires costly shape rendering and inevitably loses the internal geometric structure and spatial information.

Voxel-based methods require first preprocessing a given point cloud into voxels. Then, a voxel-based convolutional neural network is applied to extract features. Such methods can easily overcome point cloud density variations but are hampered by training costs that grow exponentially with voxel resolution. Typical works include VoxelNet \cite{VoxNet} and Minkowski Engine \cite{Minkowski-CNN}. These methods designed octree-based convolution and sparse convolution to extract local representations of point clouds, effectively reducing the consumption of GPU memory and computing costs.

Point-based methods, which directly take point clouds as input and process them in an unstructured format, have attracted increasing attention due to the absence of information loss and high training efficiency. PointNet \cite{PointNet} is a pioneering work , which proposes to model the permutation invariance of points by max-pooling point-wise features. PointNet++ \cite{PointNet++} improves PointNet by further gathering local features in a hierarchical way. DGCNN \cite{DGCNN} considers a point cloud as a graph and dynamically updates the graph to aggregate features.  Recently Transformer \cite{Transformer} based methods have been proposed as a new paradigm for processing point clouds \cite{PCT, Point-Bert, PointTransformer, LCPFormer}.

In this work, we combine point-based methods and view-based methods to achieve cross-modal information fusion.

\subsection{Unsupervised Domain Adaptation}
Unsupervised domain adaptation (UDA) has been extensively explored on images \cite{ADDA, MCD, MDD, CBST, MATHS, CycleST, CyCADA, PixelDA, NRMT}, which aims of mitigating the domain gap between source domain containing labled data and target domain containing unlabled data. These methods can generally be categorized into three categories.
1) Adversarial training \cite{ADDA, MCD, MDD, TAL-MIRN, DAFL, DSEM}, playing minimax games at the domain level between a discriminator and a generator. 
2) Style transfer \cite{CyCADA}, wherein the translation from the source domain to the target domain is directly learned using Generative Adversarial Networks \cite{GAN}.
3) Self-training with pseudo-labels \cite{CBST, CycleST, MATHS, STHPEF, AdaptiveR_id}, where partial supervision is provided to learn the distributions of the target domain.
Despite the extensive research on UDA for 2D images, the domain of 3D point clouds is still in its nascent stages, with some methods borrowed from image-based UDA. For instance, PointDAN \cite{PointDAN} is a pioneering work addressing UDA in point cloud classification by explicitly aligning local and global features across domains through domain adversarial training. ALSDA~\cite{ALSDA} introduces an automated loss function search method to address the issues of domain discriminator degeneration and cross-domain semantic mismatches in adversarial domain adaptation. GAST \cite{GAST} employs a self-training method equipped with self-paced learning \cite{SelfPaced} for point cloud UDA. GLRV \cite{GLRV} proposes a reliable voting-based method for pseudo label generation, while SD \cite{SD} employs Graph Neural Networks (GNNs) \cite{GAN} to refine pseudo-labels online during self-training. Chen et al. \cite{QBST} propose quasi-balanced self-training, dynamically adjusting the threshold to balance the proportion of pseudo-label samples for each category, thereby improving the quality of pseudo-labels.
In addition to the mainstream methods of UDA for 2D images, recent works on UDA for point clouds primarily focus on designing suitable self-supervised pretext tasks to facilitate the learning of domain-invariant features. For example, GAST \cite{GAST} proposes rotation classification and distortion localization as a self-supervised task to align features at both local and global levels. DefRec \cite{DefRec} introduces deformation-reconstruction, and Learnable-Defrec \cite{Learnable-Defrec} extends it into a learnable deformation task to further enhance performance. RS \cite{RS} shuffles and restores the input point cloud to improve discrimination. GLRV \cite{GLRV} proposes two self-supervised auxiliary tasks: scaling-up-down prediction and 3D-2D-3D projection reconstruction, along with a reliable pseudo-label voting strategy to further enhance domain adaptation. GAI \cite{GAI} employs a self-supervised task of learning geometry-aware global implicit representations for domain adaptation on point clouds.
Differentiating from the above single-modal self-supervised methods, we propose a cross-modal self-supervised task that uses 2D images to reconstruct 3D point clouds, thereby empowering the network with the ability to extract 3D geometric information from 2D images.

\subsection{2D-to-3D Knowledge Transferring}
The concept of model compression was originally introduced by Bucila et al. \cite{ModelCompression}, with the aim of transferring knowledge from a large model to a smaller one without significant performance degradation. Hinton et al. \cite{KD} systematically summarized existing knowledge distillation techniques, showcasing the effectiveness of the student-teacher strategy and response-based knowledge distillation. Recently, the transfer of 2D knowledge to 3D using view-based methods has garnered considerable attention among researchers. For instance, PointCLIP \cite{PointCLIP} directly utilized the pretrained CLIP \cite{CLIP} model for zero-shot point cloud classification via image projection. The subsequent version, PointCLIP V2 \cite{PointCLIPv2}, refined the projection strategy, resulting in a significant performance boost. ULIP \cite{ULIP, CLIP2Point} employs large multimodal models to generate detailed language descriptions of 3D objects, addressing limitations in existing 3D object datasets regarding the quality and scalability of language descriptions. PointCMD \cite{PointMCD} explores the transfer of cross-modal knowledge from multi-view 2D visual modeling to 3D geometric modeling to facilitate the understanding of the shape of the 3D point cloud. PointVST \cite{PointVST} introduces a self-supervised task that utilizes projected multi-view 2D images as self-supervised signals, enhancing the representation capabilities of point-based networks. I2P-MAE \cite{I2P-MAE} proposes a pre-training framework that leverages 2D pre-trained models to guide the learning of 3D representations. More advanced methods exploit point-pixel correspondences \cite{SPARF, VointCloud, 3Dto2DKD, P2P} between point clouds and multi-view projected images. Image2Point \cite{Image2Point} presents a kernel inflation technique that expands kernels of a 2D CNN into 3D kernels and applies them to voxel-based point cloud understanding. There is a growing interest in utilizing pre-trained Transformers for point cloud processing. PCExpert \cite{PCExpert} and EPCL \cite{EPCL} directly train high-quality point cloud models using pre-trained Transformer models as encoders. Although the Transformer pre-trained on large-scale 2D image data possesses powerful semantic representation capabilities, it lacks the ability to capture 3D information. Therefore, in this work, we follow the approach of PCExpert and EPCL, maintaining a Transformer pretrained on ImageNet \cite{imagenet} as an encoder for 3D point clouds, while also designing a self-supervised training task to reconstruct masked 3D point clouds using masked 2D images.

\section{Proposed Methods}
This section introduces the overall working mechanism and specific technical implementations of the proposed RPD. We first introduce and formulate the unsupervised domain adaptation problem on point cloud in Sec. \ref{sec: definition}.  
Then we present general formulations of deep image encoders and deep point encoders respectively in Sec. \ref{sec: teacher} and Sec. \ref{sec: student}, based on which we construct a unified online cross-modal knowledge distillation workflow in Sec. \ref{sec: distillation}. Furthermore, we introduce a novel self-supervised task to reconstruct masked point cloud patches with masked multi-view image in Sec. \ref{sec: reconstruction}. After that, self-Training strategy is described in detail in Sec. \ref{sec: self-training}. In the end, we summarize the overall loss function and training strategy in Sec. \ref{sec: overall}.

\subsection{Problem Definition}
\label{sec: definition}
Given a source domain $\mathcal{S} = \{\mathcal{P}_i^s, \mathcal{I}_i^s, y_i^s\}_{i=1}^{n_s}$ with $n_s$ labeled synthetic samples and a target domain $\mathcal{T} = \{\mathcal{P}_i^t, \mathcal{I}_i^t\}_{i=1}^{n_t}$ with $n_t$ unlabeled real samples, a semantic label space $\mathcal{Y}$ is shared between $\mathcal{S}$ and $\mathcal{T}$ (\ie $\mathcal{Y}^{s} = \mathcal{Y}^{t}$), where $\mathcal{P} \in \mathbb{R}^{N \times 3}$ represents a point cloud consisting of $N$ three-dimensional spatial coordinate points $(x,y,z)$, and $\mathcal{I} \in \mathbb{R}^{V \times W \times H}$ represents $V$ views 2D point-based projected images with a resolution of $W \times H$, and the superscripts $s$ and $t$ denote the source and target domains, respectively.
Let input space $\mathcal{X} = \{\mathcal{P}, \mathcal{I}\}$, our goal is to learn a domain-adapted mapping function $\Phi: \mathcal{X} \rightarrow \mathcal{Y}$ that can correctly classify target samples with accessing labeled source domain and unlabeled target domain. The mapping function $\Phi = \Phi^\mathcal{P} \oplus \Phi^\mathcal{I}$,  can be formulated into a cascade of a feature encoder $\Phi_\text{fea}: \mathcal{X} \rightarrow \mathbb{R}^d$ for any input $\{\mathcal{P}, \mathcal{I}\}$ and a classifier $\Phi_\text{cls}: \mathbb{R}^{d}\rightarrow [0,1]^c$ typically using fully-connected layers as follows:
\begin{equation}
\begin{aligned}
&\Phi^\mathcal{P}(\mathcal{P}) = \Phi_\text{cls}^\mathcal{P}(\bm{z}^\mathcal{P}) \circ \Phi_\text{fea}^{\mathcal{P}}(\mathcal{P}), \\
&\Phi^\mathcal{I}(\mathcal{I}) = \Phi_\text{cls}^\mathcal{I}(\bm{z}^\mathcal{I}) \circ \Phi_\text{fea}^\mathcal{I}(\mathcal{I}), \\
&{\rm logit} = \Phi^\mathcal{P}(\mathcal{P}) \oplus  \Phi^\mathcal{I}(\mathcal{I}),
\end{aligned}
\end{equation}
where $\oplus$ denotes cross-modal ensemble, $d$ denotes the dimension of the feature representation output $\bm{z} \in \mathbb{R}^d$ of $\Phi_\text{fea}(\mathcal{.})$, $c$ denotes the number of shared classes and the superscripts $\mathcal{P}$ and $\mathcal{I}$ denote the point and image modalities, respectively.

\subsection{Teacher Network for Image Modeling}
\label{sec: teacher}
Owning to the maturity of deep convolutional architectures, we can directly resort to powerful 2D models of different architectures (ResNet \cite{resnet}, ViT \cite{vit}, Clip \cite{CLIP}, \cite{effec_fusion}) for image feature fusion and extraction. 
Benefiting from the common practice of large-scale pretraining (e.g., on ImageNet \cite{imagenet} and Conceptual Captions \cite{CLIP}), the resulting 2D deep feature encoder demonstrates strong generalization ability when fine-tuned on downstream visual recognition tasks. This excellent property makes the pre-trained 2D model suitable as a teacher model for image feature extraction.
To align the input modality for 2D models, we project the input point cloud onto multiple image planes, and then encode them into multi-view 2D representations.
Specifically, given a point cloud $\mathcal{P}$, we first project it into multiple single-channel depth maps $\{\mathcal{I}_v\}_{v=1}^V \in \mathbb{R}^{V \times H \times W}$ via \textbf{Realistic Projection Pipeline} introduced by PointClip v2 \cite{PointCLIPv2}, where $V$ and $(H, W)$ denote the number of view-images and image size, respectively. 
Then the teacher image encoder take multi-view images $\{\mathcal{I}_v\}_{v=1}^V$ in parallel as input to extract image features.

In this paper, we employ a MAE \cite{MAE} pre-trained ViT \cite{vit} to encode image feature. Formally, given a single-channel depth image $\mathcal{I}_v \in \mathbb{R}^{H \times W}$, the ViT divides the image into a sequence of flattened local image patches $\{\bm{x}^{\mathcal{I}}_{v,i}\}_{i=1}^{N_{\mathcal{I}}} \in \mathbb{R}^{N_{\mathcal{I}} \times P^2}$ and used a tokenizer $\Phi_\text{emb}^{\mathcal{I}}$ (\ie Conv2D) to convert these patches into a sequence of 1-D visual token embeddings: 
\begin{equation}
\begin{aligned}
&\{\bm{z}_{v,i}^{\mathcal{I}}\}_{i=1}^{N_{\mathcal{I}}} = \Phi_\text{emb}^{\mathcal{I}} \big(\{\bm{x}^{\mathcal{I}}_{v,i}\}_{i=1}^{N_{\mathcal{I}}} \big),
\end{aligned}
\end{equation}
where $\{\bm{z}_{v,i}^{\mathcal{I}}\}_{i=1}^{N_{\mathcal{I}}} \in \mathbb{R}^{N_{\mathcal{I}} \times D_1}$, $N_{\mathcal{I}} = HW / P^2$ denotes the number of tokens, $(P, P)$ denotes the resolution of image patches, and $D_1$ is the dimension of each image token embedding. 
A learnable class token embedding $\bm{z}_\text{cls}^{\mathcal{I}}$ is prepended to the sequence of the patch embeddings. Then, the final image input representation $\mathcal{H}_{v}^{\mathcal{I}} \in \mathbb{R}^{(N_{\mathcal{I}} + 1) \times D_1}$  are calculated by summing the image patch embedding with image position embeddings $\mathcal{Z}_{\text{pos}, v}^{\mathcal{I}} \in \mathbb{R}^{(N_{\mathcal{I}} + 1) \times D_1}$:
\begin{equation}
\begin{aligned}
&\mathcal{H}_{v}^{\mathcal{I}} = [\bm{z}_\text{cls}^{\mathcal{I}}, \bm{z}_{v,1}^{\mathcal{I}}, ..., \bm{z}_{v,N_{\mathcal{I}}}^{\mathcal{I}}] + \mathcal{Z}_{\text{pos}, v}^{\mathcal{I}}
\end{aligned}
\end{equation}
Formally, the behaviours of the 2D teacher transformer module $\mathcal{M}_t$ can be formulated as follows:
\begin{equation}
\begin{aligned}
&\{\widehat{\mathcal{Z}}_{v}^{\mathcal{I}}\}_{v=1}^V = \mathcal{M}_t \big( \{\mathcal{H}_{v}^{\mathcal{I}}\}_{v=1}^V \big), \\
&\bm{\hat{z}}^{\mathcal{I}} = Concat(\{\bm{\hat{z}}_{v, 0}^{\mathcal{I}}\}_{v=1}^V), \\
&\bm{z}^{\mathcal{I}} = Proj(\bm{\hat{z}}^{\mathcal{I}}),
\end{aligned}
\end{equation}
where $\widehat{\mathcal{Z}}_{v}^{\mathcal{I}} = \{\bm{\hat{z}}_{v,i}^{\mathcal{I}}\}_{i=0}^{N_{\mathcal{I}}} \in \mathbb{R}^{(N_{\mathcal{I}} + 1) \times D_2}$ 
with subscript $v$ represents a set of view-specific image token features extracted from image $\mathcal{I}_v$, $\bm{\hat{z}}_{v, 0}^{\mathcal{I}}$ denote a view-specific class token feature, $\bm{\hat{z}}^{\mathcal{I}} \in \mathbb{R}^{VD_2}$ denotes concatenation of all view-specific class token features, $Proj$ denotes a projector based on a multi-layer perceptron (MLP) with three fully connected layers, and $\bm{z}^{\mathcal{I}} \in \mathbb{R}^d$ denotes the final feature representation of the image modality input. By default $V = 10, P = 16, H = W = 224, N_{\mathcal{I}} = 196, D_1 = 768, D_2 = 512$.


\subsection{Student Network for 3D Point Cloud Modeling}
\label{sec: student}
Collecting and labeling 3D shape models is costly and time-consuming, resulting in the current 3D community still lacking large-scale and richly-annotated datasets comparable to those in the 2D field (\ie \cite{imagenet, coco}). Limited by the insufficiency of training data, the parameters of mainstream point cloud networks (\ie \cite{PointCNN, PointNet, PointNet++, DGCNN}) are actually small to alleviate overfitting. This makes these point cloud networks poorly scalable and unsuitable for ``pretrain-and-finetune". We believe that Transformer-based models are inherently well-suited for learning robust semantic patterns in point clouds due to their ability to capture the topological configurations of local geometries.
Before the standard transformer is applied to the point cloud field, there are some transformer layers (\cite{PCT, PointTransformer}) specifically designed for point cloud processing. Pioneered by PointBERT \cite{Point-Bert}, the standard transformer has been applied to point cloud tasks. 

Following \cite{Point-Bert}, we sample $N_\mathcal{P}$ centroids using Furthest Point Sampling (FPS). To each of these centroids, we assign $k$ neighbouring points by conducting a $k$-Nearest Neighbour (KNN) search. Thereby, we obtain $N_\mathcal{P}$ local geometric patches $\{\bm{x}^{\mathcal{P}}_{i}\}_{i=1}^{N_{\mathcal{P}}} \in \mathbb{R}^{N_{\mathcal{P}} \times (k+1) \times 3}$, where each geometric patch $\bm{x}^{\mathcal{P}}_{i}$ consists of a centroid $\bm{x}^{\mathcal{P}}_{i, 0}$ and its $k$ neighboring point $\{\bm{x}^{\mathcal{P}}_{i, j}\}_{j=1}^{k}$, \ie $\bm{x}^{\mathcal{P}}_{i} = \{\bm{x}^{\mathcal{P}}_{i, j}\}_{j=0}^{k}$. These patches are subsequently fed into tokenizer $\Phi_\text{emb}^{\mathcal{P}}$  (mini-DGCNN \cite{DGCNN}) to obtain patch token embeddings: 
\begin{equation}
\begin{aligned}
&\{\bm{z}_{i}^{\mathcal{P}}\}_{i=1}^{N_{\mathcal{P}}} = \Phi_\text{emb}^{\mathcal{P}} \big(\{\bm{x}^{\mathcal{P}}_{i}\}_{i=1}^{N_{\mathcal{P}}} \big),
\end{aligned}
\end{equation}
where $\{\bm{z}_{i}^{\mathcal{P}}\}_{i=1}^{N_{\mathcal{P}}} \in \mathbb{R}^{N_{\mathcal{P}} \times D_1}$, $N_{\mathcal{P}}$ denotes the number of geometric tokens and $D_1$ denotes the feature dimension. 
Similarly, a learnable class token embedding $\bm{z}_\text{cls}^{\mathcal{P}}$ is prepended to the sequence of the patch embeddings. Then, the final point cloud input representation $\mathcal{H}^{\mathcal{P}} \in \mathbb{R}^{(N_{\mathcal{P}} + 1) \times D_1}$  are calculated by summing the geometric patch embedding with position embeddings $\mathcal{Z}_{\text{pos}}^{\mathcal{P}} \in \mathbb{R}^{(N_{\mathcal{P}} + 1) \times D_1}$:
\begin{equation}
\begin{aligned}
&\mathcal{H}^{\mathcal{P}} = [\bm{z}_\text{cls}^{\mathcal{P}}, \bm{z}_{1}^{\mathcal{P}}, ..., \bm{z}_{N_{\mathcal{P}}}^{\mathcal{P}}] + \mathcal{Z}_{\text{pos}}^{\mathcal{P}}
\end{aligned}
\end{equation}
Formally, the 3D student transformer module $\mathcal{M}_s$ consumes $\mathcal{H}^{\mathcal{P}}$ and outputs high-dimensional feature representation $\bm{\hat{z}}^{\mathcal{P}}$, which can be described as:
\begin{equation}
\begin{aligned}
&\widehat{\mathcal{Z}}^{\mathcal{P}} = \mathcal{M}_s \big(\mathcal{H}^{\mathcal{P}} \big), \\
&\bm{z}^{\mathcal{P}} = Proj(\bm{\hat{z}}_0^{\mathcal{P}}),
\end{aligned}
\end{equation}
where $\widehat{\mathcal{Z}}^{\mathcal{P}} = \{\bm{\hat{z}}_i^{\mathcal{P}}\}_{i=0}^{N_{\mathcal{P}}} \in \mathbb{R}^{(N_{\mathcal{P}} + 1) \times D_2}$ denotes the embedded point cloud token features, $\bm{\hat{z}}_0^{\mathcal{P}}$ denotes the embedded class token feature, $Proj$ is a three-layer MLP, and $\bm{z}^{\mathcal{P}} \in \mathbb{R}^d$ denotes the final feature representation of the point cloud modality input. By default $N_{\mathcal{P}}=27, k = 128, D_1 = 768, D_2 = 512$.


\subsection{Online Cross-Modal Knowledge Distillation}
\label{sec: distillation}
Here, we aim to explore how the knowledge from pre-trained 2D Transformer models can be utilized for 3D feature representation learning.
On the one hand, the 2D teacher model pre-trained on large-scale data sets (\ie ImageNet \cite{imagenet}) has strong capabilities to learn high-quality representation, \ie robust and generalizable features, stemming from their ability to capture long-range correlations across local patches.
This prior knowledge of modeling the relationships between local parts is ideal for guiding 3D models to capture the topology of local geometries, eliminating the need for pre-training on large 3D geometry datasets.
On the other hand, it is evident that the transformer modules of both the teacher model ($\mathcal{M}_t$) and the student model ($\mathcal{M}_s$) are structurally identical, consisting of a series of layer normalization (LN), multi-head self-attention (MSA) and multi-layer perceptron (MLP) layers. The only difference lies in the tokenizer during feature extraction.
Therefore, distilling relational priors from a 2D pre-trained model to a 3D model is highly feasible without requiring additional complex designs.

To harness the relational priors ingrained in pre-trained 2D teacher model for 3D representation learning, we propose a strategy of parameter sharing and online knowledge distillation for 2D-to-3D knowledge transfer.
First, we share a parameter-frozen pre-trained transformer module between the 2D teacher model ($\mathcal{M}_t$) and the 3D student model ($\mathcal{M}s$), while keeping the image tokenizer parameters ($\Phi_\text{emb}^{\mathcal{I}}$) in the 2D teacher model frozen during training.
Second, we distill the teacher model's semantic knowledge into the student model by imposing the following cross-modal alignment constraint:
\begin{equation}
\begin{aligned}
&\mathcal{L}_\text{kd} =  D_\text{KL}\big(\Phi_\text{cls}^\mathcal{P}(\bm{z}^\mathcal{P})||  \Phi_\text{cls}^\mathcal{I}(\bm{z}^\mathcal{I})\big),
\end{aligned}
\end{equation}
where $D_{KL}$ denotes KL-divergence loss function, $\Phi_\text{cls}^\mathcal{P}$ and $\Phi_\text{cls}^\mathcal{I}$ represent classifiers of 2D teacher model and 3D student model respectively. More details aboout online distillation process are given in Algorithm \ref{distll_algorithm}.

\begin{algorithm}[htb]
\SetKwInOut{Input}{Input}
\SetKwInOut{Output}{Output}
\SetKwInOut{Parameter}{Parameter}
\SetKwInOut{Initialization}{Initialization}
\label{distll_algorithm}
\caption{Online Distillation Process}
\Input{~\\
\noindent labeled source data $\mathcal{S} = \{\mathcal{P}_i^s, \mathcal{I}_i^s, y_i^s\}_{i=1}^{n_s}$;\\
\noindent unlabeled target data $\mathcal{T} = \{\mathcal{P}_i^t, \mathcal{I}_i^t\}_{i=1}^{n_t}$;\\
\noindent student network $\Phi^\mathcal{P}(\mathcal{P}) = \Phi_\text{cls}^\mathcal{P}(\bm{z}^\mathcal{P}) \circ \Phi_\text{fea}^{\mathcal{P}}(\mathcal{P})$;\\ 
\noindent teacher network $\Phi^\mathcal{I}(\mathcal{I}) = \Phi_\text{cls}^\mathcal{I}(\bm{z}^\mathcal{I}) \circ \Phi_\text{fea}^\mathcal{I}(\mathcal{I})$;\\
\noindent decoder $\Phi_\text{dec}^\mathcal{P}$;\\
\noindent number of epochs $E$;\\
}
        
\Output{~\\
\noindent $\Phi^\mathcal{P}$ and $\Phi^\mathcal{I}$
}

\Initialization {~\\
\noindent initialize $\mathcal{M}_t$ and $\mathcal{M}_s$ with pre-trained Vit and fix the parameters of first nine blocks;}

        \For{$e\leftarrow 1$ \KwTo $E$}{
            \For{$(\mathcal{P}_i^s, \mathcal{I}_i^s, y_i^s), (\mathcal{P}_i^t, \mathcal{I}_i^t)$ in $(\mathcal{S}, \mathcal{T})$}{
                \eIf{$e\ \%\ 10 < 5$}{
                    $\min_{\Phi^\mathcal{P}, \Phi^\mathcal{I}} \mathcal{L}_\text{cls}^s$ with $(\mathcal{P}_i^s, y_i^s)$;\\
                    $\min_{\Phi^\mathcal{P}, \Phi^\mathcal{I}} \mathcal{L}_\text{kd}$ with $\mathcal{P}_i^s$ and $\mathcal{P}_i^t$;\\
                    $\min_{\Phi_\text{dec}^\mathcal{P}} \mathcal{L}_\text{emd}$ with $\mathcal{P}_i^s$ and $\mathcal{P}_i^t$;\\
                    }{
                    $\min_{\Phi^\mathcal{P}} \mathcal{L}_\text{cls}^s$ with $(\mathcal{P}_i^s, y_i^s)$;\\
                    $\min_{\Phi^\mathcal{P}} \mathcal{L}_\text{kd}$ with $\mathcal{P}_i^s$ and $\mathcal{P}_i^t$;\\
                    $\min_{\Phi_\text{dec}^\mathcal{P}} \mathcal{L}_\text{emd}$ with $\mathcal{P}_i^s$ and $\mathcal{P}_i^t$;\\
                    }
            }
        }
\end{algorithm}

\subsection{Masked Point Cloud Reconstruction}
\label{sec: reconstruction}
Transferring the knowledge of 2D pre-trained models for 3D feature representation learning lacks awareness of 3D geometric information.
Motivated by SiamMAE \cite{SiameseMAE}, we design a self-supervision task that reconstructs masked point cloud patches with corresponding masked multi-view image features to capture 3D geometric information of point clouds. 
Specially, given a sequence of $N_{\mathcal{P}}$ tokens embeddings of point cloud local patches 
$\{\bm{\hat{z}}_{i}^{\mathcal{P}}\}_{i=1}^{N_{\mathcal{P}}}$, 
we randomly mask these token embeddings with high mask ratio ($85\%$). A set of learnable mask embeddings 
$\{\bm{m}_{i}^{\mathcal{P}}\}_{i=1}^{M_{\mathcal{P}}}$, where 
$M_{\mathcal{P}} = \lfloor0.85 \times N_{\mathcal{P}} \rfloor$, 
initialized with Gaussian distribution $N(0, 0.02)$ are used to replace the masked positions and are set as the query inputs of the joint decoder $\Phi_\text{dec}$. The unmasked token embeddings of point cloud patches are denoted as 
$\{\bm{r}_i^{\mathcal{P}}\}_{i=1}^{R_{\mathcal{P}}}$ where 
$R_{\mathcal{P}} = N_{\mathcal{P}} - M_{\mathcal{P}}$.
Then, the corresponding $N_{\mathcal{I}} \times V$ image tokens embeddings 
$\{\widehat{\mathcal{Z}}_{v}^{\mathcal{I}} - \widehat{\mathcal{Z}}_{v,0}^{\mathcal{I}} \}_{v=1}^V$ 
are set as the key and value input of the joint decoder to reconstruct the masked point cloud patches, where $\widehat{\mathcal{Z}}_{v,0}^{\mathcal{I}} = \{\bm{\hat{z}}_{v, 0}^{\mathcal{I}}\}$ denote the set of view-specific class token feature. 
Considering redundant information and computation efficiency, we randomly drop the image token embeddings with high drop ratio ($85\%$), the remaining image token embeddings are represented as 
$\{\bm{r}_i^{\mathcal{I}}\}_{i=1}^{R_{\mathcal{I}}}$ where 
$R_{\mathcal{I}} = \lfloor0.15 \times N_{\mathcal{I}} \times V \rfloor$.  
We believe that asymmetric masking/dropping can create a challenging self-supervised learning task while encouraging the network to learn 3D geometric information.

The joint decoder has two layers and each layer consists of a multi-head cross-attention (MCA) and a multi-head self-attention layer (MSA). A fully connected linear layer (FCL) is used to project the output of the decoder to the reconstructed point cloud. Formally, the behaviours of the decoder $\Phi_\text{dec}$ can be formulated as follows:
\begin{equation}
\begin{aligned}
&\mathcal{F}_0 = \{\bm{m}_{i}^{\mathcal{P}}\}_{i=1}^{M_{\mathcal{P}}} 
\cup 
\{\bm{r}_{i}^{\mathcal{P}}\}_{i=1}^{R_{\mathcal{P}}},\\
&\mathcal{F}_1 = \text{MSA} \big( \text{MCA} \big(
\mathcal{F}_0,
\{\bm{r}_{i}^{\mathcal{I}}\}_{i=1}^{R_{\mathcal{I}}}
\big) \big), \\
&\mathcal{F}_2 = \text{MSA} \big( \text{MCA} \big(
\mathcal{F}_1,
\{\bm{r}_{i}^{\mathcal{I}}\}_{i=1}^{R_{\mathcal{I}}}
\big)\big), \\
&\mathcal{R} = \text{FCL}(\mathcal{F}_2),
\end{aligned}
\end{equation}
where $\mathcal{R}$ denotes the reconstracted point cloud. The distance between $\mathcal{R}$ and the original point cloud $\mathcal{P}$ is calculated using Earth Mover's Distance (EMD) distance. Thereby, the loss function for the reconstruction task is defined as:
\begin{equation}
\begin{aligned}
&\mathcal{L}_\text{emd} =  D_\text{EMD}(\mathcal{R} || \mathcal{P}),
\end{aligned}
\end{equation}
where $D_\text{EMD}$ denotes the EMD distance measure function.

\subsection{Self-Training}
\label{sec: self-training}
Before adaptation, both 2D teacher model and 3D student model take labeled source domain data 
(\ie $\{\mathcal{P}_i^s, \mathcal{I}_i^s, y_i^s\}_{i=1}^{n_s}$) as input for supervised learning:
\begin{equation}
\begin{aligned}
\mathcal{L}_\text{cls}^s = - \frac{1}{n_s} \sum_{i=1}^{n_s} \sum_{c=1}^{C} {\rm I}[c = y_i^s] \log \big(\Phi^\mathcal{P}(\mathcal{P}_i^s)_c \Phi^\mathcal{I}(\mathcal{I}_i^s)_c\big),
\end{aligned}
\end{equation}
where $\Phi^\mathcal{P}(\mathcal{P}_i^s)_c$ and $\Phi^\mathcal{I}(\mathcal{I}_i^s)_c$ denote the predicted probabilities of the $c$-th class of the teacher model and student model respectively, and $\rm I[\cdot]$ is an indicator function.

For adaptation, self-paced self-training (SPST) is a popular strategy to align the two domains by generating pseudo-labels in the target domain according to highly confident predictions. Follow these works \cite{GAST, GLRV, GAI, QBST}, we also utilize SPST strategy to further reduce domain shift. The objective of self-paced learning based self-training is depicted as:
\begin{align}
\label{EqnTrgSelfLearn}
\begin{aligned}
 \mathcal{L}_\text{cls}^t = - \frac{1}{\widehat{n}_t} \sum_{i=1}^{\widehat{n}_t} \left( \sum_{c=1}^C \widehat{y}_{i,c}^t \log \big( \Phi^\mathcal{P}(\mathcal{P}_i^t)_c \Phi^\mathcal{I}(\mathcal{I}_i^t)_c \big) + \gamma |\widehat{\bm{y}}_i^t|_1 \right),
\end{aligned}
\end{align}
where $\widehat{n}_t$ denotes the number of the pseudo labeled samples in target domain, $\widehat{\bm{ y}}_i^t$ is the predicted pseudo label one-hot vector for a target instance $\mathcal{P}_i^t$, $\widehat{y}_{i,c}^t$ is its $c$-th element, and $\gamma$ is a hyper-parameter controls the number of selected target samples, \ie the larger $\gamma$, the more samples. We can simply convert $\gamma$ into the prediction confidence threshold $\theta = \exp(- \gamma)$. The generic pseudo-label generation strategy can be simplified to the following form when all network parameters are fixed:
\begin{eqnarray}
\label{LabelAssign}
\widehat{y}^t_{i,c} =\! \left\{
    \begin{aligned}
    & 1, \:\: {\rm if} \: c = \arg\max_{c} p(c|{\rm logit}_i) \: \text{\&} \ p(c|{\rm logit}_i) > \theta, \\
    & 0, \:\: {\rm otherwise},
    \end{aligned}
    \right.
\end{eqnarray}
where ${\rm logit}_i = Avg(\Phi^\mathcal{P}(\mathcal{P}_i^t), \Phi^\mathcal{I}(\mathcal{I}_i^t))$. We adopt a threshold $\theta$ that gradually increases with self-paced rounds evolve, \ie each round increases by a constant $\epsilon$.


\subsection{Overall Loss}
\label{sec: overall}
The framework of our approach is illustrated in Fig. \ref{fig:model}. The overall training loss of our method is:
\begin{equation}
\begin{aligned}
\mathcal{L} =  \mathcal{L}_\text{kd} + \alpha \mathcal{L}_\text{emd} + \beta \mathcal{L}_\text{cls}^s + \eta \mathcal{L}_\text{cls}^t,
\end{aligned}
\end{equation}
where $\alpha$, $\beta$ and $\eta$ are hyper-parameters used to balance the weights between methods. We follow \cite{GAI, GAST, GLRV, QBST} to apply a two-stage optimization for training the models. During the first stage of model training, we mainly rely on the first three loss terms to ensure better completion of the adaptation process. Once the initial training is completed, we use the trained teacher and student models together to generate pseudo labels for the target domain samples and perform the self-training.
\section{Experiments}
\subsection{Datasets}
\noindent \textbf{PointDA-10.} The PointDA-10 \cite{PointDAN} is a popular UDA dataset designed for point cloud classification, which consists of subsets of three datasets: ShapeNet, ModelNet40 and ScanNet. These sub-datasets share the same ten categories like bathtub, bed, and bookshelf. In particular, ShapeNet-10(\textbf{S}) is the subset of ShapeNet dataset and contains 17,378 training and 2,492 testing point cloud extracted from synthetic 3D CAD models. Similarly, ModelNet-10(\textbf{M}) consists of 4,183 training and 856 testing samples taken from the synthetic dataset ModelNet40, but the shape of the point cloud exhibits variations from the same class samples in ShapeNet. ScanNet-10(\textbf{S*}) is sampled from ScanNet and contains 6,110 training samples and 1,769 testing samples, respectively. It is the only real dataset of scanned real-world indoor scenes. Due to errors in the registration process and occlusions, the point clouds in ScanNet-10 suffer from issues of noise and sparseness, making classification more challenging. With the three sub-datasets, we can evaluate our method in six different UDA settings including Simulation-to-Reality, Reality-to-Simulation and Simulation-to-Simulation scenarios.

\noindent \textbf{Sim-to-Real.} The Sim-to-Real \cite{MetaSets} dataset is a fairly new benchmark for the problem of 3D domain generalization (3DDG), which collects object point clouds of 11 shared classes from ModelNet40 \cite{Modelnet} and ScanObjectNN \cite{ScanObjectNN}, and 9 shared classes from ShapeNet \cite{Shapenet} and ScanObjectNN \cite{ScanObjectNN}. This benchmark consists of four subsets: ModelNet-11 (\textbf{M11}), ScanObjectNN-11 (\textbf{SO*11}), ShapeNet-9 (\textbf{S9}) and ScanObjectNN-9 (\textbf{SO*9}). Among them, \textbf{M11} consists of 4,844 training and 972 testing point clouds, \textbf{SO*11} includes 1,915 training and 475 testing point clouds, \textbf{S9} consists of 1,9904 training and 1,995 testing point clouds, \textbf{SO*9} includes 1,602 training and 400 testing point clouds. 
Following \cite{GLRV}, we conduct two types of Simulation-to-Reality adaptation scenarios: \textbf{M11} $\rightarrow$ \textbf{SO*11} and \textbf{S9} $\rightarrow$ \textbf{SO*9}.

\subsection{Implementation Details}
For our RPD, we adopt mini-DGCNN \cite{DGCNN} as \textit{3D Tokenizer} $\Phi_\text{emb}^{\mathcal{P}}$ which is a standard DGCNN with half the number of layers. The \textit{2D Tokenizer} $\Phi_\text{emb}^{\mathcal{I}}$ is a 2D convolution layer with a convolution kernel size equal to the image patch size.
We adopt a standard vision transformer 
as the backbone to extract relationships across patch tokens from images and point clouds. The transformer module is initialized by MAE \cite{MAE} pre-trained ViT-B/16 \cite{vit} and we only train the last three blocks to balance accuracy and efficiency. 
The \textit{Category Classifier} $\Phi_\text{cls}^{\mathcal{I}}$ and $\Phi_\text{cls}^{\mathcal{P}}$ are based on a multi-layer perceptron (MLP) with three fully connected layers.
The \textit{Joint Decoder} $\Phi_\text{dec}$ for self-supervised reconstruction has two layers and each layer consists of a multi-head cross-attention (MCA) and a multi-head self-attention (MSA) layer, followed by a fully connected linear (FCL) projection layer.
By default, the hyper-parameters of $\alpha, \beta $ and $\eta$ are empirically set to 1, 1 and 1 respectively. During training, the Adam optimizer \cite{adam} is utilized with the initial learning rate 0.0001 and the epoch-wise cosine annealing learning rate scheduler. Dropout of 0.5 and batch normalization were adaptively applied after the convolution layers and the hidden layers. The training batch size is set to 32. More training details are provided in Table \ref{tab:configs}.
During self-spaced self-training (SPST), the initial threshold $\theta$ and the increment constant $\epsilon$ are empirically set to 0.8 and 0.05 and the training contains 10 rounds, with 5 epochs in each round. 
For simulation-to-reality scenarios, some specific data augmentation strategies were adopted, such as jittering, randomly dropping holes and rotation.

\noindent \textbf{Transformer Configurations:}
We extract relatiobships between image and point cloud using the standard ViT \cite{vit} architecture, which comprises 12 layers of 12 attention heads and an embedding dimensions of 768. Only the last three layers are trained to balance accuracy and efficiency. The decoder network has 2 layers, each equipped with a multi-head cross-attention (MCA) and a multi-head self-attention (MSA)layer. The number of attention heads and embedding dimensions are set to 16 and 512, respectively.

\begin{table*}[ht]
\centering
\caption{Training configurations for 6 settings in PointDA-10 \cite{PointDAN} and 2 settings in Sim-to-Real \cite{MetaSets}. The R, J, D in augmentation denote rotation, jittering and randomly dropping holes respectively.}
\renewcommand{\arraystretch}{1.15}
\resizebox{0.9\linewidth}{!}{
\begin{tabular}{l|cccccccc}
\hline
\textbf{Config} & M$\rightarrow$S & M$\rightarrow$S* & S$\rightarrow$M & S$\rightarrow$S* & S*$\rightarrow$M & S*$\rightarrow$S & S9$\rightarrow$SO*9 & M11$\rightarrow$SO*11\\
\hline
optimizer & Adam & Adam & Adam & Adam & Adam & Adam & Adam & Adam \\
base learning rate & 1e-4 & 1e-4 & 1e-4 & 1e-4 & 1e-4 & 1e-4 & 1e-4 & 1e-4 \\
weight decay & 5e-5 & 5e-5 & 5e-4 & 5e-4 & 5e-5 & 5e-5 & 5e-5 & 5e-5\\
dropout & 0.5 & 0.5 & 0.5 & 0.5 & 0.5 & 0.5 & 0.5 & 0.5\\
training epochs & 400 & 400 & 200 & 200 & 200 & 200 & 400 & 400\\
label smoothing & 0 & 0 & 0.3 & 0.3 & 0 & 0 & 0 & 0 \\
augmentation & R & R, J, D & R & R, J, D & R, J & R, J & R, J, D & R, J, D\\
\hline
\end{tabular}}
\label{tab:configs}
\end{table*}

\begin{table*}[ht]
\caption{Classification accuracy (\%) averaged over 3 seeds ($\pm$ SEM) on the PointDA-10 dataset. M: ModelNet-10; S: ShapeNet-10; S*: ScanNet-10. We compare with the state-of-the-art 3D UDA methods and our method achieves best performance. {$\dagger$} denotes experiments without using 3 seeds.The best performance is highlight in bold}
\label{tab:compare_sota}
\centering
\renewcommand{\arraystretch}{1.15}
\resizebox{0.9\linewidth}{!}{
\begin{tabular}
{lc|ccccccc}
\hline
Methods & SPST & M$\rightarrow$S & M$\rightarrow$S* & S$\rightarrow$M & S$\rightarrow$S* & S*$\rightarrow$M & S*$\rightarrow$S & Avg  \\
\hline
w/o Adapt 
& & 83.3$\pm$0.7 & 43.8$\pm$2.3 & 75.5$\pm$1.8 & 42.5$\pm$1.4 & 63.8$\pm$3.9 & 64.2$\pm$0.8 & 62.2$\pm$1.8 \\
\hline
PointDAN \cite{PointDAN} 
& & 83.9$\pm$0.3 & 44.8$\pm$1.4 & 63.3$\pm$1.1 & 45.7$\pm$0.7 & 43.6$\pm$2.0 & 56.4$\pm$1.5 & 56.3$\pm$1.2 \\
RS \cite{RS} 
& & 79.9$\pm$0.8 & 46.7$\pm$4.8 & 75.2$\pm$2.0 & 51.4$\pm$3.9 & 71.8$\pm$2.3 & 71.2$\pm$2.8 & 66.0$\pm$1.6 \\
DefRec+PCM \cite{DefRec} 
& & 81.7$\pm$0.6 & 51.8$\pm$0.3 & 78.6$\pm$0.7 & 54.5$\pm$0.3 & 73.7$\pm$1.6 & 71.1$\pm$1.4 & 68.6$\pm$0.8 \\
Learnable-DefRec{$^\dagger$} \cite{Learnable-Defrec} 
& & 82.8$\pm$0.0 & 56.3$\pm$0.0 & 81.7$\pm$0.0 & 54.8$\pm$0.0 & 72.9$\pm$0.0 & 71.7$\pm$0.0 & 70.0$\pm$0.0 \\
GLRV\cite{GLRV}
& \checkmark & 85.4$\pm$0.4 & 60.4$\pm$0.4 & 78.8$\pm$0.6 & 57.7$\pm$0.4 & 77.8$\pm$1.1 & 76.2$\pm$0.6 & 72.7$\pm$0.6 \\
\multirow{2}{*}{GAST \cite{GAST}} 
& & 83.9$\pm$0.2 & 56.7$\pm$0.3 & 76.4$\pm$0.2 & 55.0$\pm$0.2 & 73.4$\pm$0.3 & 72.2$\pm$0.2 & 69.5$\pm$0.2 \\
& \checkmark & 84.8$\pm$0.1 & 59.8$\pm$0.2 & 80.8$\pm$0.6 & 56.7$\pm$0.2 & 81.1$\pm$0.8 & 74.9$\pm$0.5 & 73.0$\pm$0.4 \\
\multirow{2}{*}{GAI \cite{GAI}} 
& & 85.8$\pm$0.3 & 55.3$\pm$0.3 & 77.2$\pm$0.4 & 55.4$\pm$0.5 & 73.8$\pm$0.6 & 72.4$\pm$1.0 & 70.0$\pm$0.5 \\
& \checkmark & 86.2$\pm$0.2 & 58.6$\pm$0.1 & 81.4$\pm$0.4 & 56.9$\pm$0.2 & 81.5$\pm$0.5 & 74.4$\pm$0.6 & 73.2$\pm$0.3 \\
SD{$^\dagger$} \cite{SD}
& \checkmark & 83.9$\pm$0.0 & 61.1$\pm$0.0 & 80.3$\pm$0.0 & 58.9$\pm$0.0 & 85.5$\pm$0.0 & 80.9$\pm$0.0 & 75.1$\pm$0.0 \\
\hline
\multirow{2}{*}{Ours} 
& & 81.9$\pm$0.3 & 64.4$\pm$0.5 & 82.8$\pm$0.4 & 59.0$\pm$0.3 & 77.1$\pm$0.8 & 76.4$\pm$0.6 & 73.6$\pm$0.5 \\
& \checkmark & \textbf{86.3$\pm$0.3} & \textbf{64.9$\pm$0.2} & \textbf{88.7$\pm$0.1} & \textbf{61.1$\pm$0.1} & \textbf{86.2$\pm$0.9} & \textbf{81.2$\pm$0.3} & \textbf{78.0$\pm$0.3} \\
\hline

\hline
\end{tabular}}
\end{table*}

\subsection{Comparison with the State-of-the-art Methods}
We compare our RPD with recent state-of-the-art point-based UDA methods including Domain Adversarial Neural Network (\textbf{PointDAN}) \cite{PointDAN}, Reconstruction Space Network (\textbf{RS}) \cite{RS}, Deformation Reconstruction Network with Point Cloud Mixup (\textbf{DefRec+PCM}) \cite{DefRec}, Learnable Deformation Reconstruction Network (\textbf{Learnable-DefRec}) \cite{Learnable-Defrec}, Global-Local structure modeling and Reliable Voted pseudo label method (\textbf{GLRV}) \cite{GLRV}, Geometry-Aware Self-Training (\textbf{GAST}) \cite{GAST}, Geometry-Aware Implicits (\textbf{GAI}) \cite{GAI}, Self-Distillation (\textbf{SD}) \cite{SD}. The \textbf{w/o Adapt} method means training the DGCNN network with only labeled source samples and is evaluated as reference of the lower performance bounds.

We report in Tab. \ref{tab:compare_sota}  the comparisons between our proposed RPD and other UDA methods on PointDA-10. As can be seen, our method surpasses all baselines by a large margin in 6 settings. The average classification accuracy of the RPD outperforms the current SOTA method SD \cite{SD} by 2.9\%. Also, the RPD achieves a remarkable enhancement over SD in the Simulation-to-Reality settings of \textbf{M$\rightarrow$S*} (+3.8 \%) and \textbf{S$\rightarrow$S*} (+2.2 \%), which are the most challenging yet realistic tasks. This observations verify the capability of our RPD to effectively capture semantic information from point clouds.

For Sim-to-Real dataset, we compare our method with meta-learning method, \ie MetaSets \cite{MetaSets}, Point-based domain adaptation methods, \ie PointDAN \cite{PointDAN} and GLRV \cite{GLRV}. We report the mean accuracy and standard error with three seeds in Table \ref{tab:sim2real}. Our method outperforms both point-based domain adaptation and meta-learning methods, achieving a new state-of-the-art.

\begin{table*}[ht]
\caption{Ablation study on each component of our method. Experiments are conducted on PointDA-10 dataset.}
\label{tab:ablation_pointda}
\centering
\renewcommand{\arraystretch}{1.15}
\resizebox{0.9\linewidth}{!}{
\begin{tabular}
{l|ccc|ccccccc}
\hline
 & OCKD & MPCR & SPST & M$\rightarrow$S & M$\rightarrow$S* & S$\rightarrow$M & S$\rightarrow$S* & S*$\rightarrow$M & S*$\rightarrow$S & Avg  \\
\hline
PointNet \cite{PointNet} & & & & 80.5 & 41.6 & 75.8 & 40.0 & 60.5 & 63.6 & 60.3 \\
DGCNN \cite{DGCNN} & & & & 83.3 & 43.8 & 75.5 & 42.5 & 63.8 & 64.2 & 62.2 \\
\hline
\multirow{6}{*}{Ours} & & & & 82.1 & 58.7 & 74.2 & 52.8 & 72.7 & 70.7 & 68.5 \\
& \checkmark & & & 82.0 & 62.6 & 75.2 & 58.3 & 74.1 & 71.0 & 70.5 \\
& & \checkmark & & 82.5 & 62.2 & 77.2 & 55.1 & 73.7 & 73.9 & 70.8 \\
& \checkmark & \checkmark & & 81.9 & 64.4 & 82.8 & 59.0 & 77.1 & 76.4 & 73.6 \\
\cline{2-11}
& & & \checkmark & 82.4 & 59.0 & 82.0 & 56.5 & 80.0 & 78.9 & 73.1 \\
& \checkmark & & \checkmark & 83.7 & 62.9 & 85.9 & 61.1 & 84.2 & 79.3 & 76.2 \\
& & \checkmark & \checkmark & 85.5 & 63.2 & 82.2 & 57.7 & 80.7 & 79.8 & 74.9 \\
& \checkmark & \checkmark & \checkmark & \textbf{86.3} & \textbf{64.9} & \textbf{88.7} & \textbf{61.1} & \textbf{86.2} & \textbf{81.2} & \textbf{78.0} \\

\hline

\hline
\end{tabular}}
\end{table*}

\begin{table}[htbp]
  \caption{Classification accuracy (\%) averaged over 3 seeds ($\pm$ SEM) on the Sim-to-Real dataset. M11: ModelNet-11; SO*11: ScanObjectNN-11; S9: ShapeNet-9; SO*9: ScanObjectNN-9.}
  \label{tab:sim2real}
  \centering
  \renewcommand{\arraystretch}{1.15}
  \resizebox{0.9\linewidth}{!}{
      \begin{tabular}{lc|cc}
      \hline
      Methods & SPST & M11$\rightarrow$SO*11 & S9$\rightarrow$SO*9 \\
      \hline
      w/o Adaptation & & 61.68$\pm$1.26 & 57.42$\pm$1.01 \\
     \hline
      PointDAN \cite{PointDAN} & & 63.32$\pm$0.85 & 54.95$\pm$0.87 \\
      MetaSets \cite{MetaSets} & & 72.42$\pm$0.21 & 60.92$\pm$0.76 \\
      GLRV \cite{GLRV} & \checkmark & 75.16$\pm$0.34 & 62.46$\pm$0.55 \\
      \hline
      \multirow{2}{*}{Ours} 
      & & 74.43$\pm$0.54 & 63.25$\pm$0.50 \\
      & \checkmark & \textbf{77.05}$\pm$0.42 &  \textbf{67.50}$\pm$0.50\\
      \hline
      
      \hline
      \end{tabular}
      }
\end{table}

\begin{table}[ht]
\caption{Ablation study on each component of our method. Experiments are conducted on Sim-to-Real dataset.}
\label{tab:ablation_sim2real}
\centering
\renewcommand{\arraystretch}{1.15}
\resizebox{0.9\linewidth}{!}{
\begin{tabular}
{ccc|cc}
\hline
 OCKD & MPCR & SPST & M11$\rightarrow$SO*11 & S9$\rightarrow$SO*9 \\
\hline
& & & 69.12 & 60.25\\
\checkmark & & & 71.24 & 61.50 \\
& \checkmark & & 70.53 & 60.75 \\
\checkmark & \checkmark & & 73.47 & 63.25\\
\cline{1-5}
& & \checkmark & 72.14 & 61.75 \\
\checkmark & & \checkmark & 74.19 & 64.50 \\
& \checkmark & \checkmark & 73.32 & 63.50\\
\checkmark & \checkmark & \checkmark & \textbf{77.05} & \textbf{67.50} \\

\hline

\hline
\end{tabular}}
\end{table}



\subsection{Ablation Studies}
To validate the effectiveness of our proposed method, we conducted various ablation studies on the six settings of PointDA-10 and two settings of Sim-to-Real. We utilized a MAE pre-trained Vision Transformer to extract features and introduced three key components for adaptation: an online cross-model knowledge distillation method (OCKD), a mask point cloud reconstruction component (MPCR), and a self-paced self-training strategy (SPST). The results are summarized in Tab. \ref{tab:ablation_pointda} and Tab. \ref{tab:ablation_sim2real}.

For PointDA-10, the first three rows in Tab. \ref{tab:ablation_pointda} respectively show the results of using PointNet \cite{PointNet}, DGCNN \cite{DGCNN}, and our proposed method as the backbone network without adaptation. It is evident that our baseline exhibits significantly better performance than PointNet \cite{PointNet} and DGCNN \cite{DGCNN} in 4 out of 6 settings, highlighting the superior generalization of transformer models pre-trained on large-scale image datasets over traditional 3D networks. By comparing the fourth row and the third row in Tab. \ref{tab:ablation_pointda}, we observe that OCKD achieves better scores across all settings than the baseline, indicating that the point cloud branch has acquired abundant semantic information, consequently enhancing its generalization capability. Furthermore, the fifth row shows that the inclusion of the mask point cloud reconstruction module improves the model's ability to capture geometric information, resulting in better classification accuracy. Moreover, a significant improvement is observed on average by using OCKD and MPCK components together. The Simulation-to-Reality settings achieve competitive results even without SPST, surpassing the performance of the previous SOTA model. Additionally, accuracy improves in all six settings after adding the SPST method, indicating its effectiveness across all datasets. Finally, in the last row, we report the results obtained by combining all components, and our method achieves the best result compared to the recent SOTA method SD \cite{SD}. For Sim-to-Real, the results are shown in Tab. \ref{tab:ablation_sim2real}, yielding similar conclusions, which once again verifies the effectiveness of our proposed method.

\begin{figure*}[ht]
    \centering
    \subfigure[w/o Adapt: M $\rightarrow$ S*]{
        \label{Fig.cofmt_m_sup.1}
        \begin{minipage}[b]{0.23\textwidth}
            \includegraphics[width=1\textwidth]{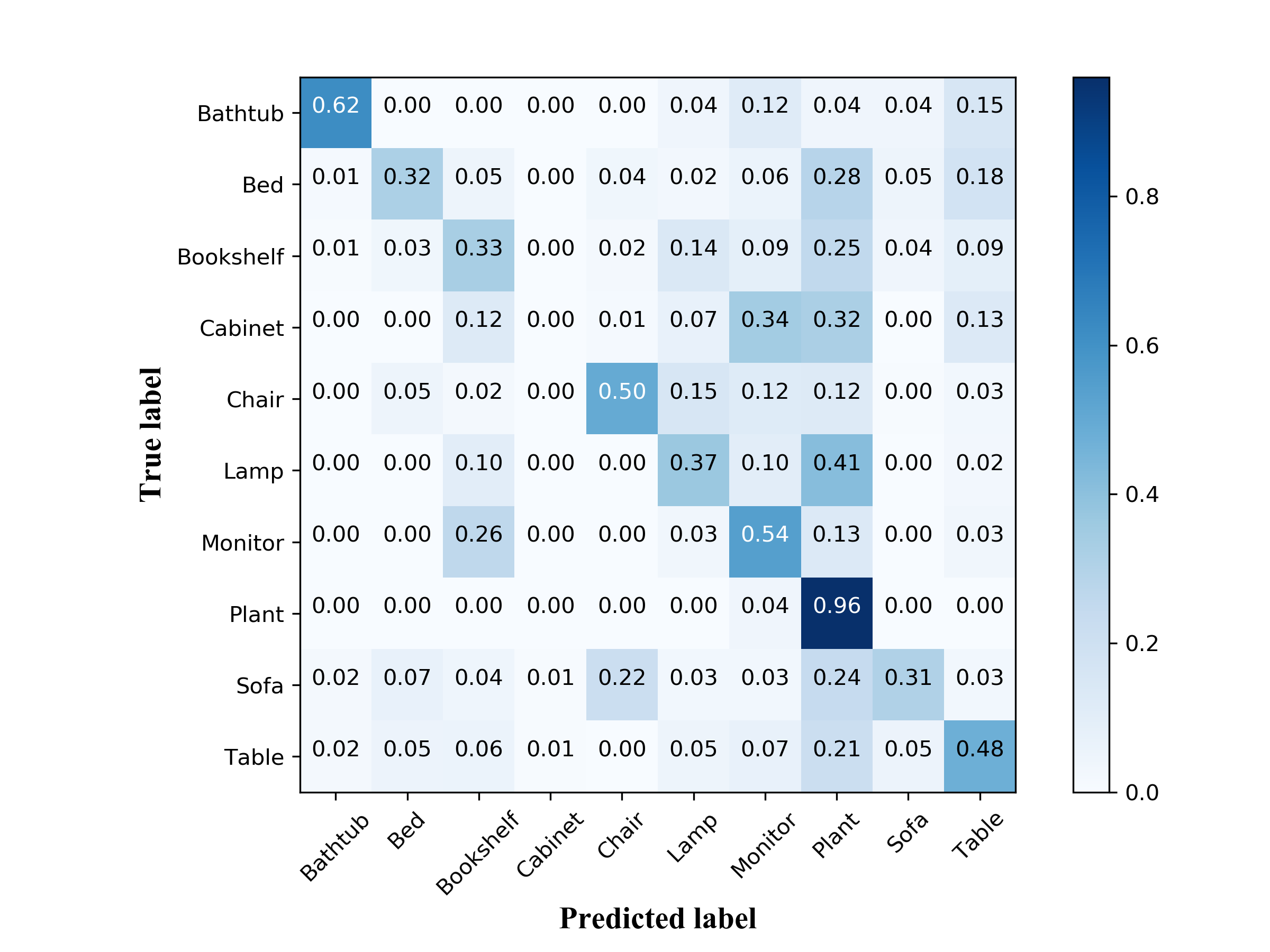}
        \end{minipage}} 
    \subfigure[Ours: M $\rightarrow$ S*]{
        \label{Fig.cofmt_m2r.2}
        \begin{minipage}[b]{0.23\textwidth}
            \includegraphics[width=1\textwidth]{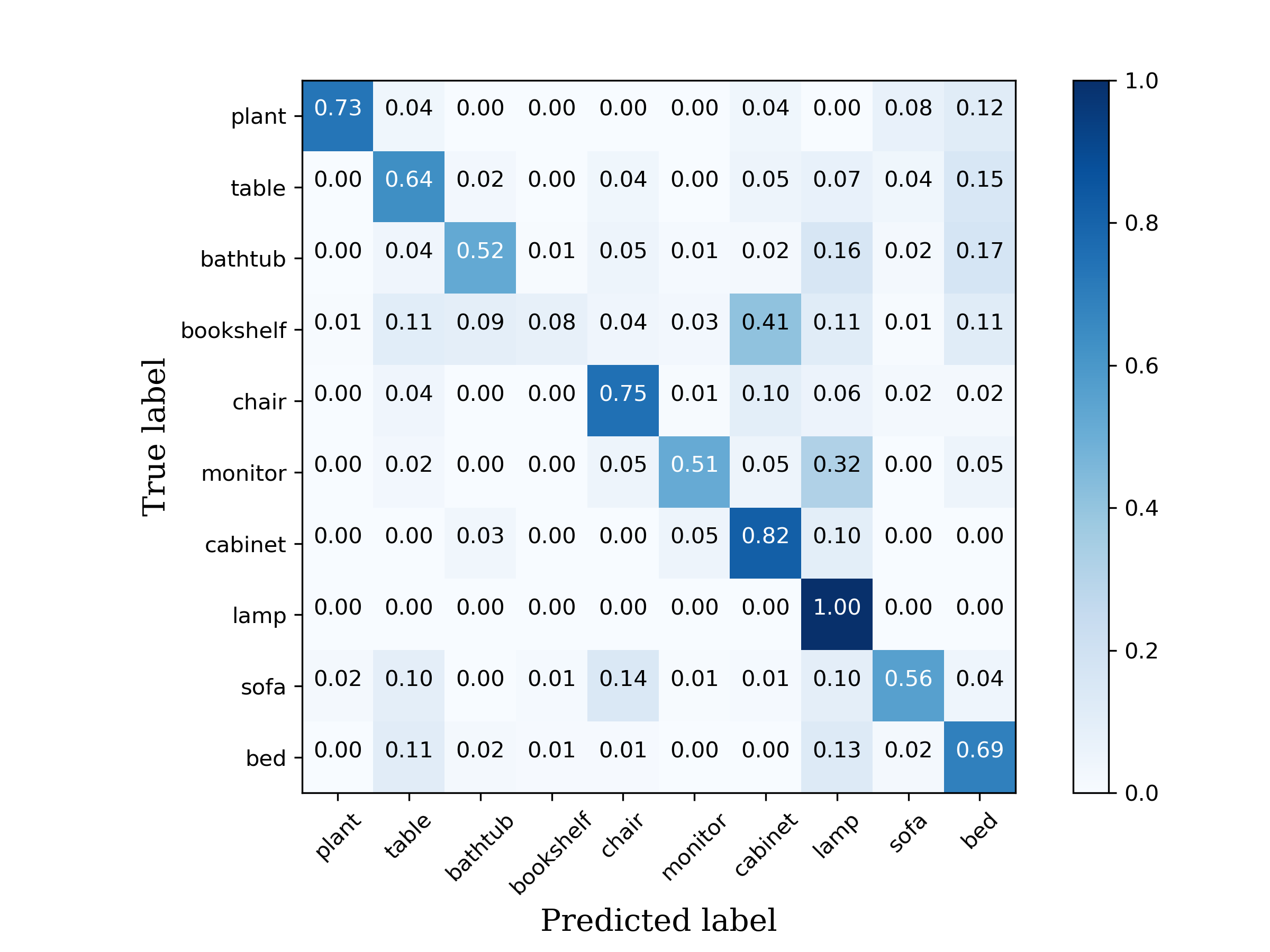}
        \end{minipage}}
    \subfigure[w/o Adapt: S $\rightarrow$ S*]{
        \label{Fig.cofmt_s_sup.3}
        \begin{minipage}[b]{0.23\textwidth}
            \includegraphics[width=1\textwidth]{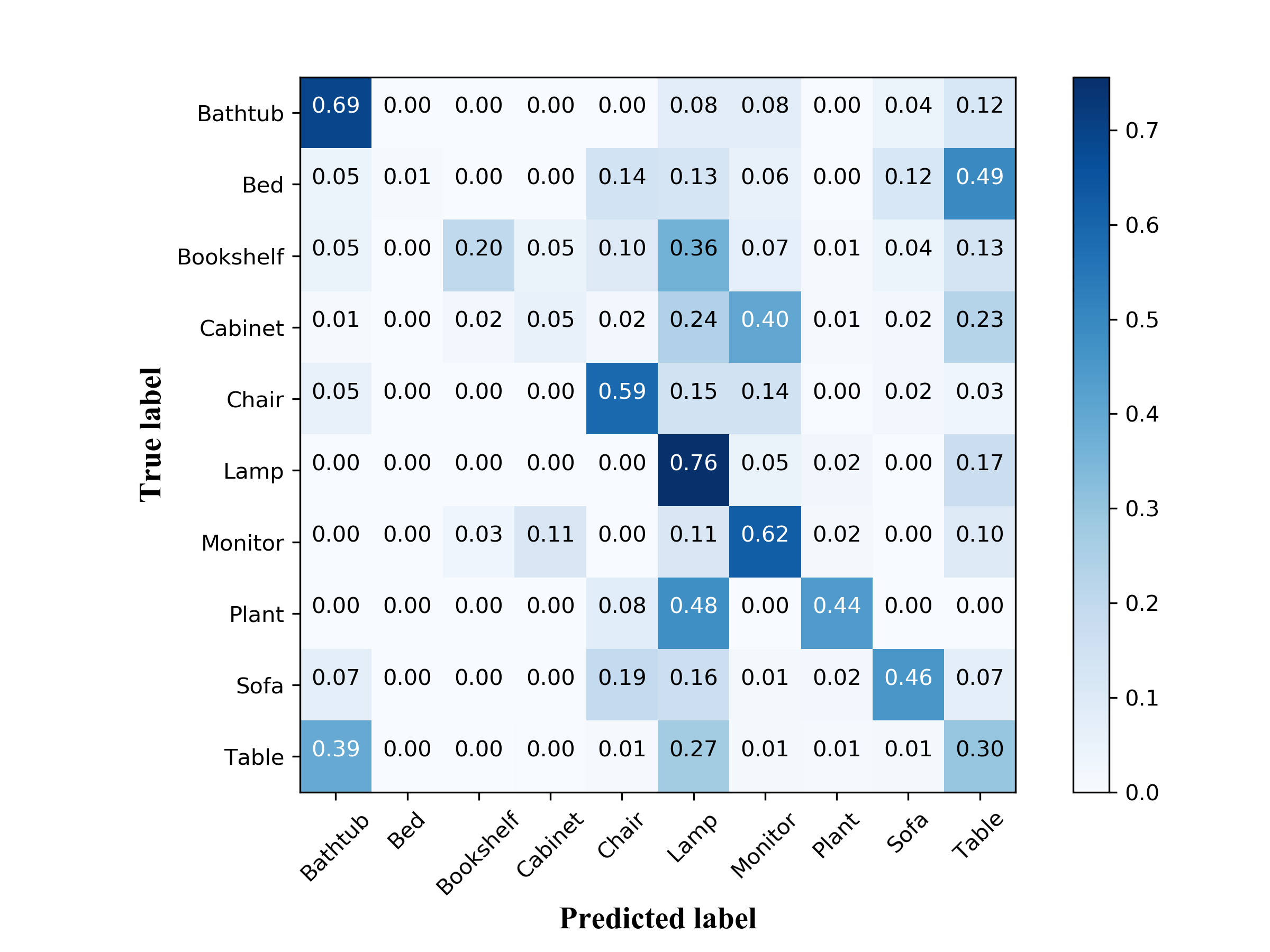}
        \end{minipage}}
    \subfigure[Ours: S $\rightarrow$ S*]{
        \label{Fig.cofmt_s2r.4}
        \begin{minipage}[b]{0.23\textwidth}
            \includegraphics[width=1\textwidth]{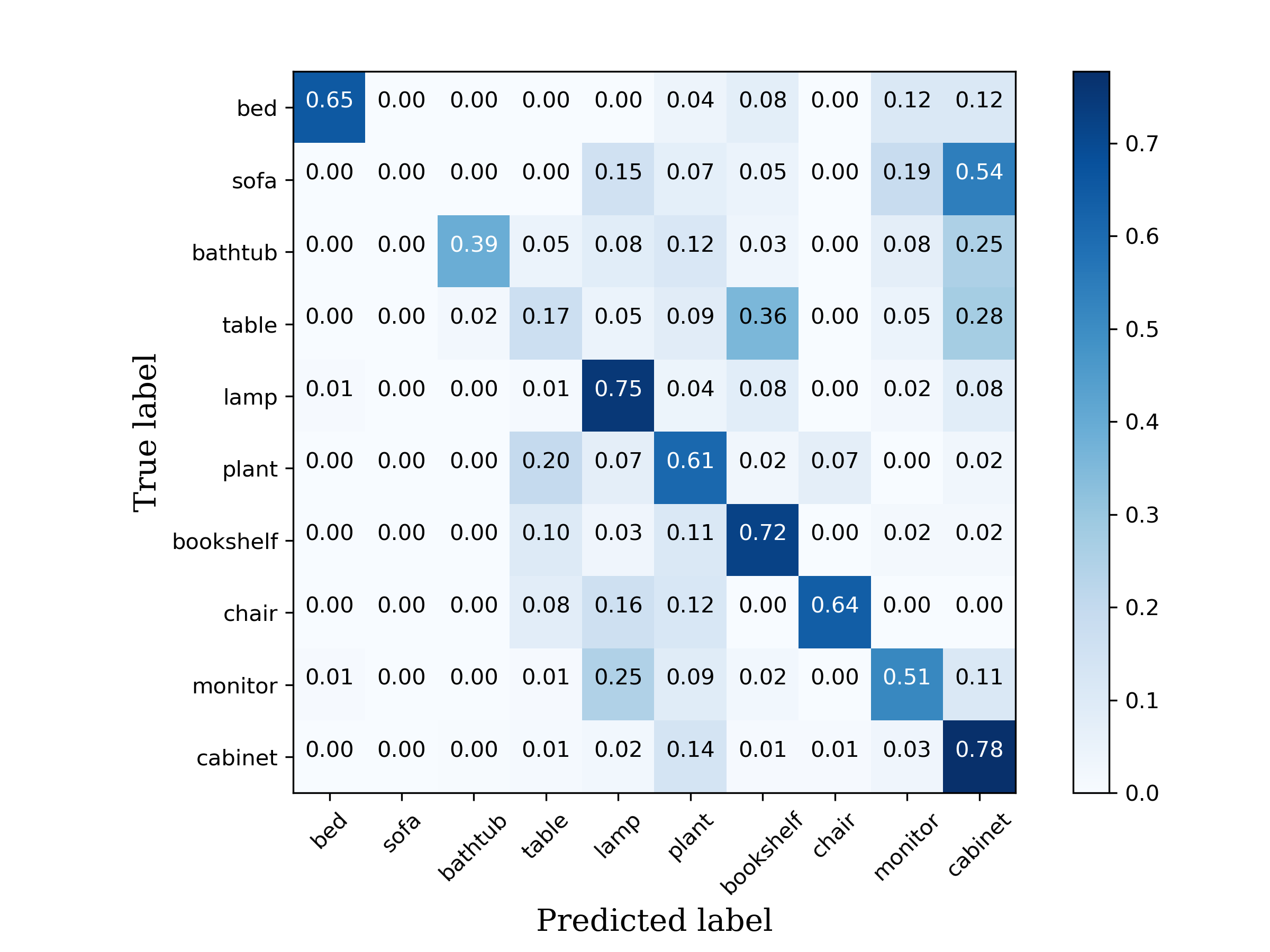}
        \end{minipage}}   
    \subfigure[w/o Adapt: M11 $\rightarrow$ SO*11]{
        \label{Fig.cofmt_m11_sup.1}
        \begin{minipage}[b]{0.23\textwidth}
            \includegraphics[width=1\textwidth]{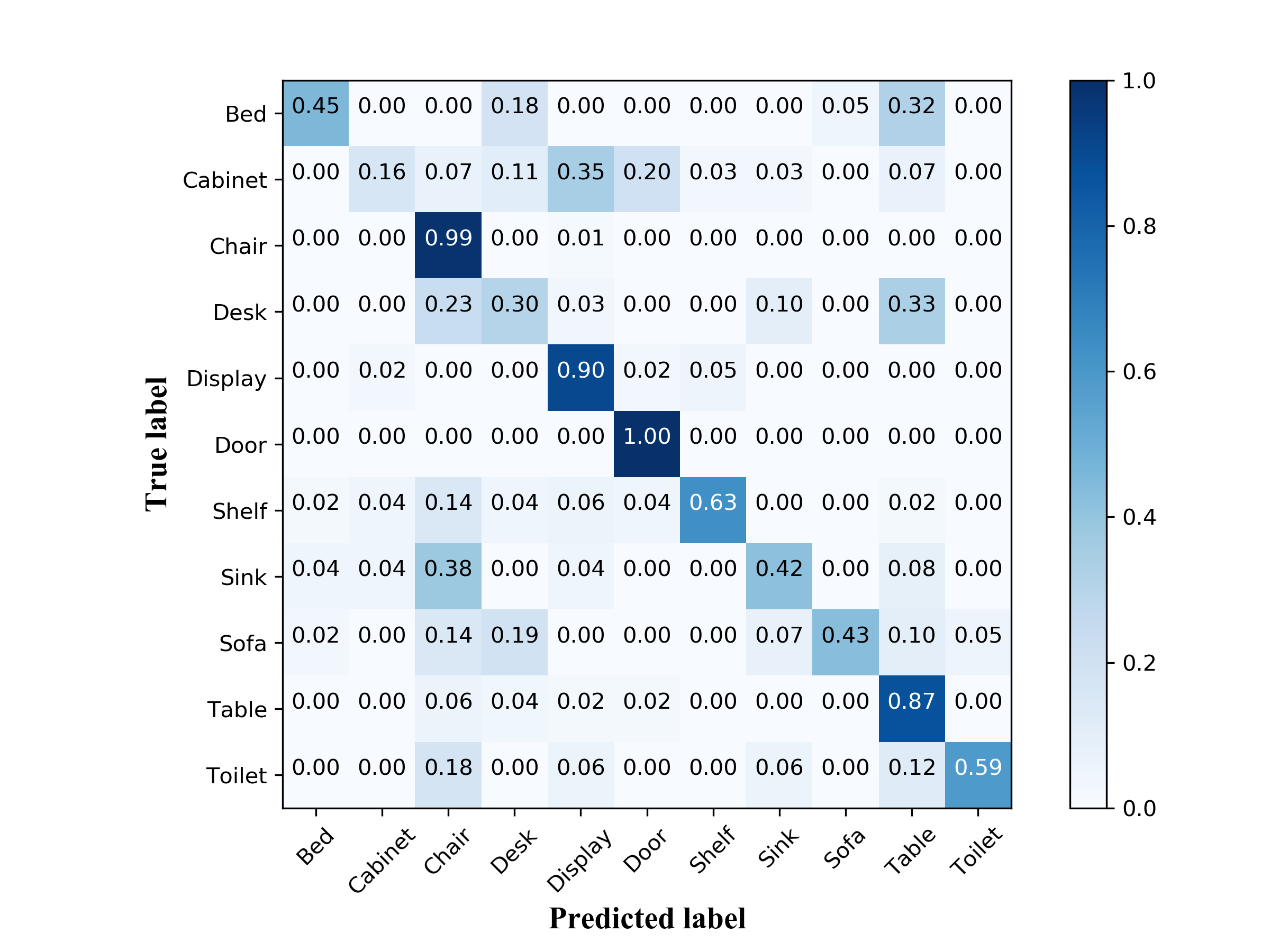}
        \end{minipage}}
    \subfigure[Ours: M11 $\rightarrow$ SO*11]{
        \label{Fig.cofmt_m112r11.2}
        \begin{minipage}[b]{0.23\textwidth}
            \includegraphics[width=1\textwidth]{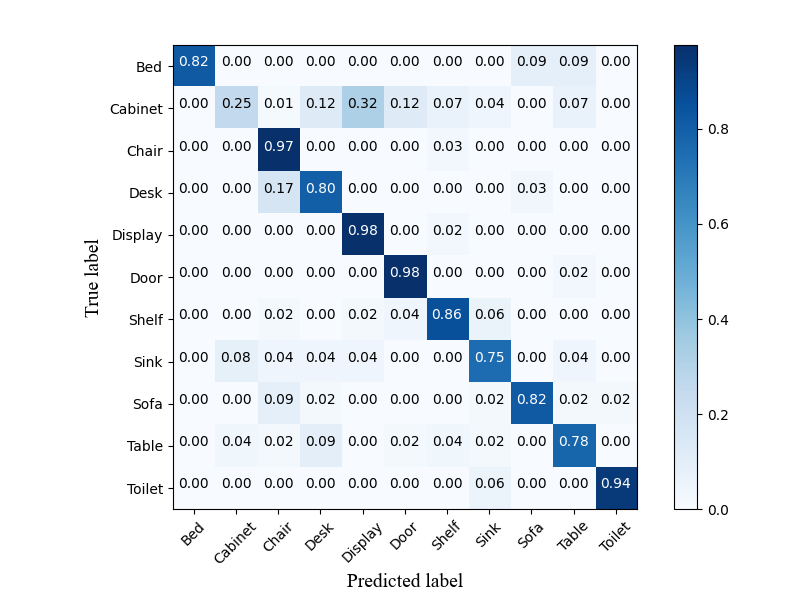}
        \end{minipage}}    
    \subfigure[w/o Adapt: S9 $\rightarrow$ SO*9]{
        \label{Fig.cofmt_s9_sup.3}
        \begin{minipage}[b]{0.23\textwidth}
            \includegraphics[width=1\textwidth]{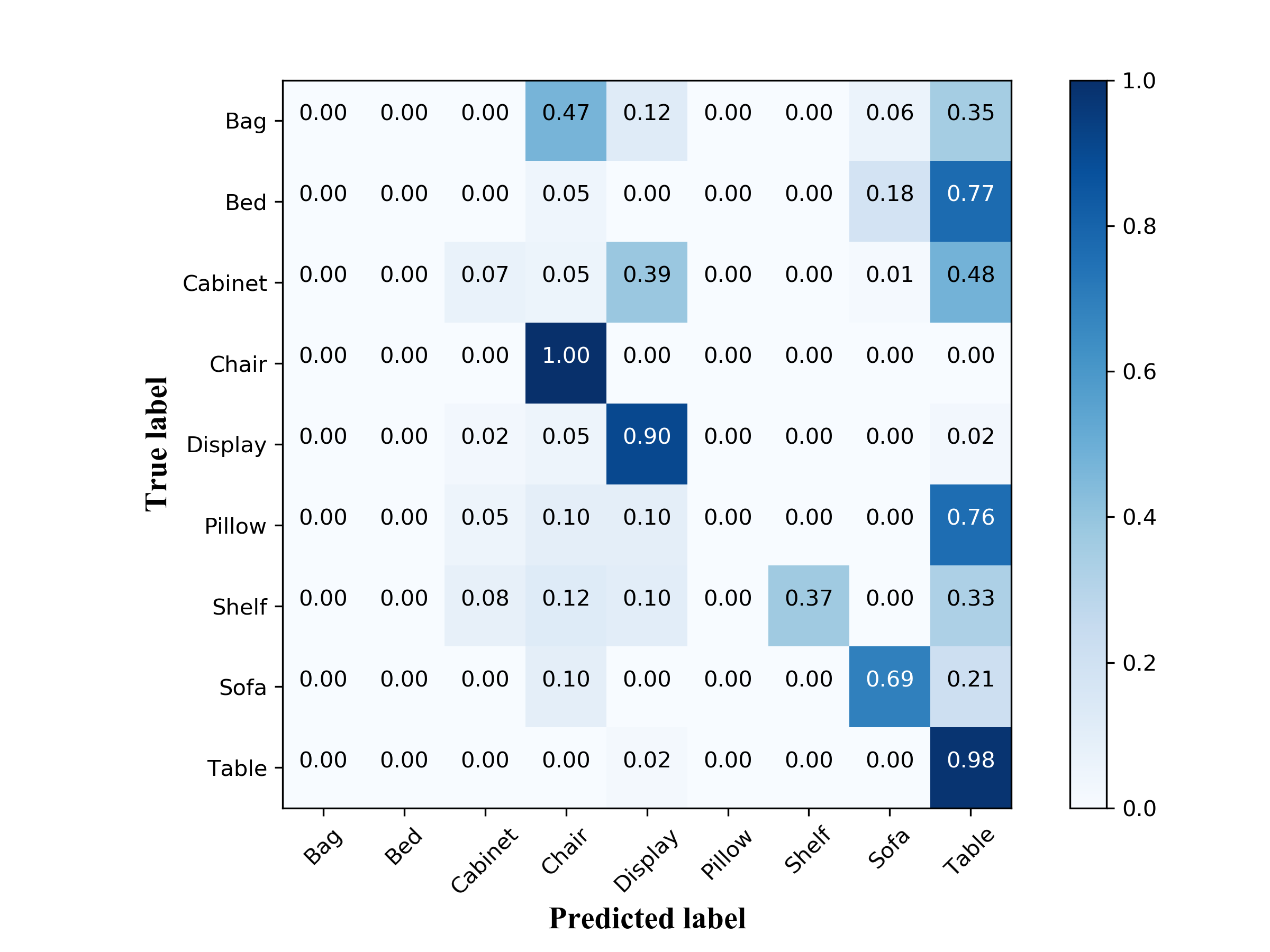}
        \end{minipage}}
    \subfigure[Ours: S9 $\rightarrow$ SO*9]{
        \label{Fig.cofmt_s92r9.4}
        \begin{minipage}[b]{0.23\textwidth}
            \includegraphics[width=1\textwidth]{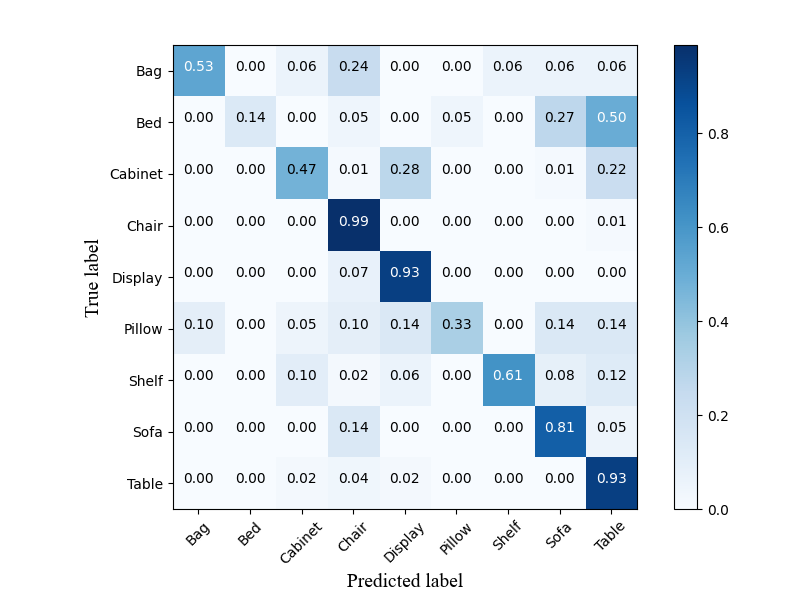}
        \end{minipage}}
    \caption{Confusion matrices of classifying testing samples on target domain under four simulation-to-reality scenarios of \textbf{M} $\rightarrow$ \textbf{S*},  \textbf{S} $\rightarrow$ \textbf{S*}, \textbf{M11} $\rightarrow$ \textbf{SO*11}, and \textbf{S9} $\rightarrow$ \textbf{SO*9}.} 
    \label{Fig.confusion_matrix}
\end{figure*}

\begin{figure}[ht]
  \centering
   \includegraphics[width=0.95\linewidth]{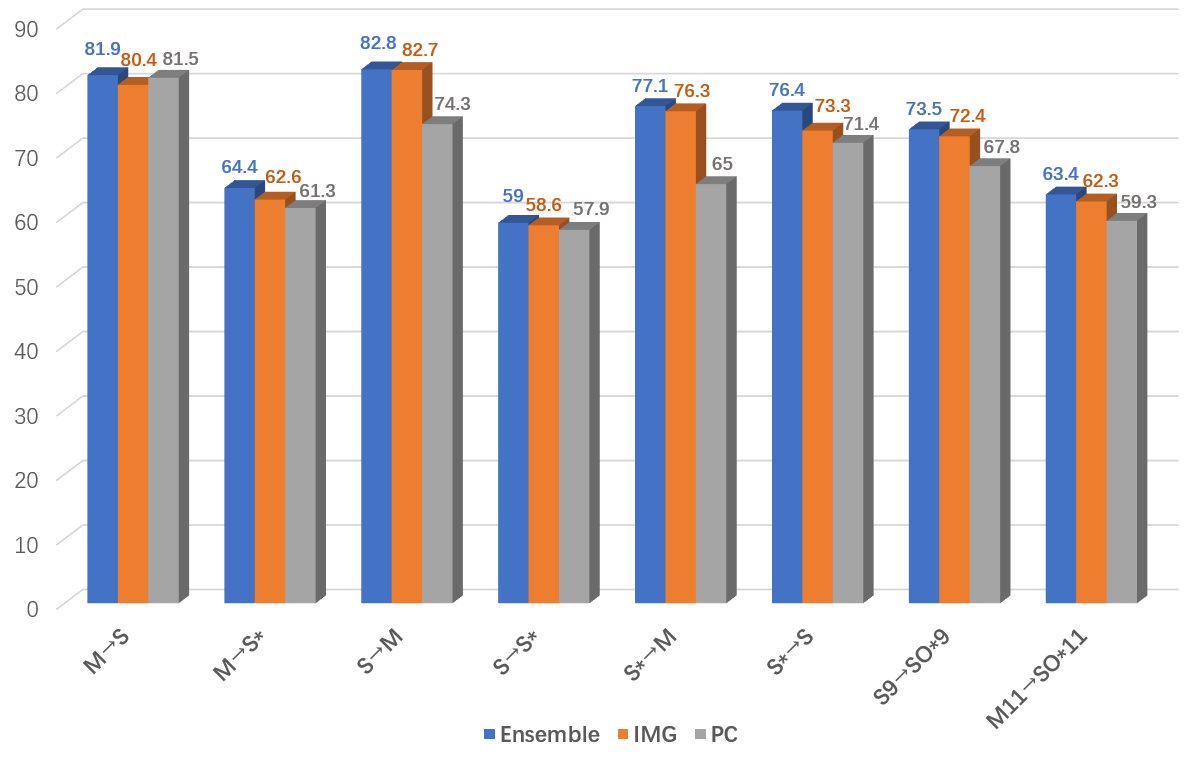}
   \caption{Illustration of cross-modal knowledge fusion.}
   \label{fig:ensemble}
\end{figure}

We also investigate the influence of the cross-modal knowledge fusion strategy. For shape classification, we directly fuse the prediction by linear interpolation, namely, adding the classification logits of 2D teacher and 3D student models element-wisely. This simple yet effective design produces the ensemble for two types of knowledge: the 3D geometric information captured by self-supervised learning masked point cloud reconstruction, and the robust semantics from the trained 2D Transformer-based models. We believe that these two kinds of knowledge have certain complementary qualities.
As shown in Fig. \ref{fig:ensemble}, cross-modal knowledge fusion strategy consistently improve the cross-domain generalization on all settings of PointDA-10 and Sim-to-Real.

\begin{figure}[ht]
    \centering
    \subfigure[M $\rightarrow$ S]{
        \label{Fig.rec_m2s}
        \begin{minipage}[b]{0.23\textwidth}
            \includegraphics[width=1\textwidth]{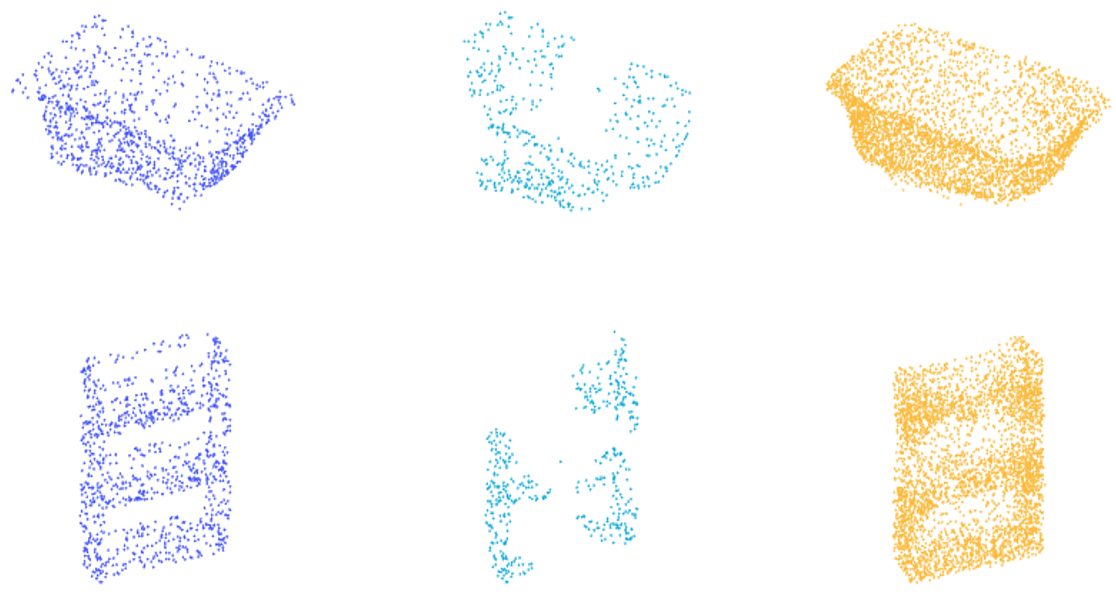}
        \end{minipage}} 
    \subfigure[M $\rightarrow$ S*]{
        \label{Fig.rec_m2r}
        \begin{minipage}[b]{0.23\textwidth}
            \includegraphics[width=1\textwidth]{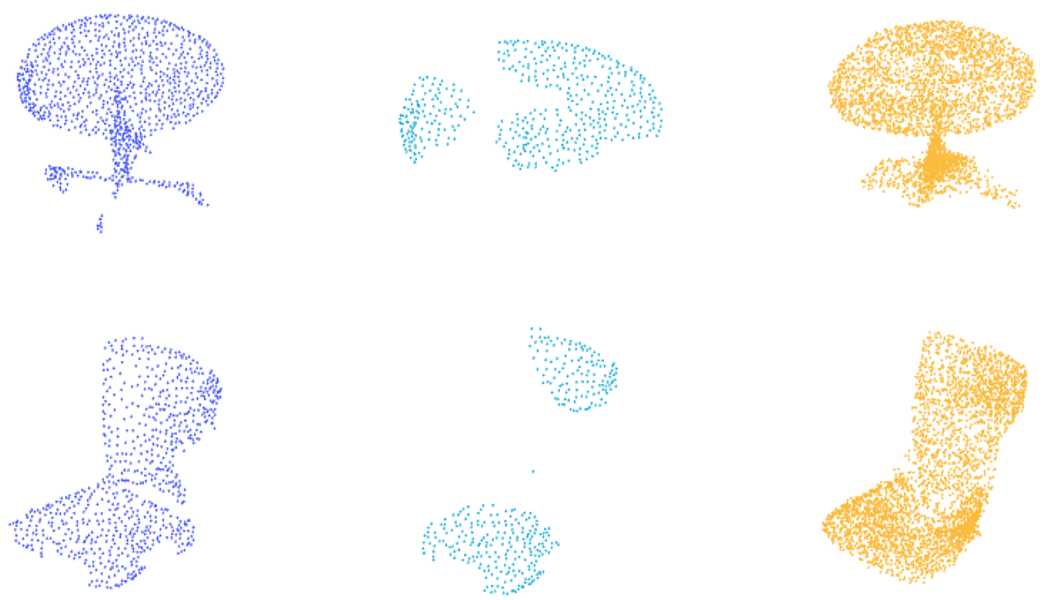}
        \end{minipage}}
    \subfigure[S $\rightarrow$ M]{
        \label{Fig.rec_s2m}
        \begin{minipage}[b]{0.23\textwidth}
            \includegraphics[width=1\textwidth]{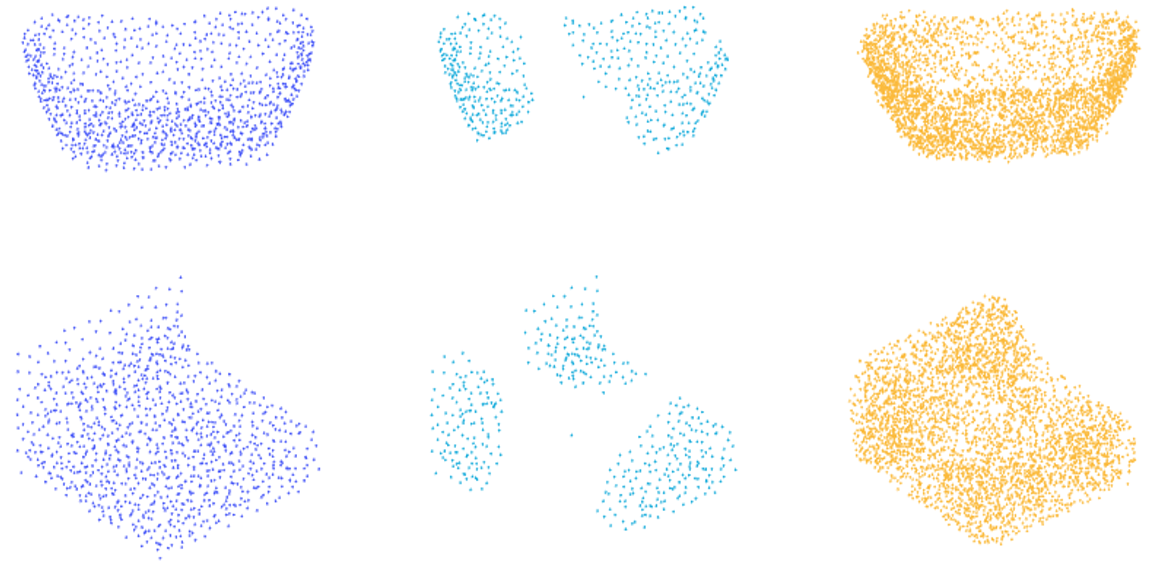}
        \end{minipage}}
    \subfigure[S $\rightarrow$ S*]{
        \label{Fig.rec_s2r}
        \begin{minipage}[b]{0.23\textwidth}
            \includegraphics[width=1\textwidth]{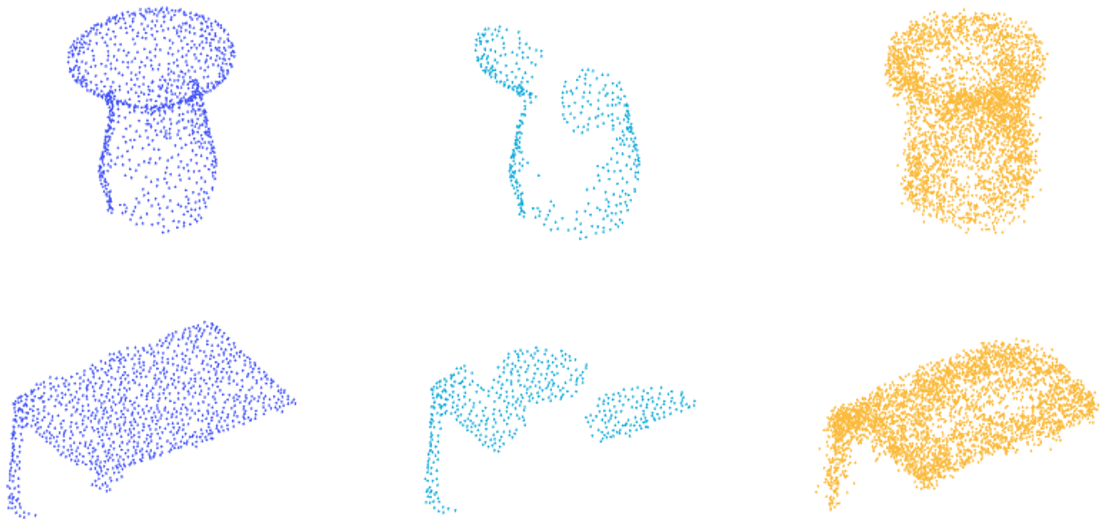}
        \end{minipage}}   
    \subfigure[S* $\rightarrow$ M]{
        \label{Fig.rec_r2m}
        \begin{minipage}[b]{0.23\textwidth}
            \includegraphics[width=1\textwidth]{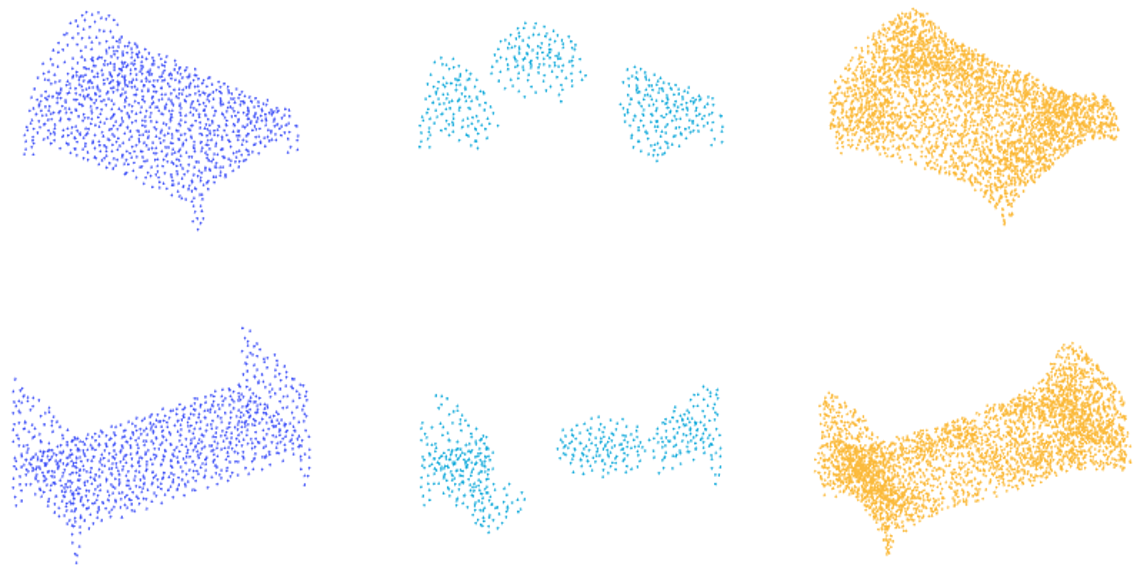}
        \end{minipage}}
    \subfigure[S* $\rightarrow$ S]{
        \label{Fig.rec_r2s}
        \begin{minipage}[b]{0.23\textwidth}
            \includegraphics[width=1\textwidth]{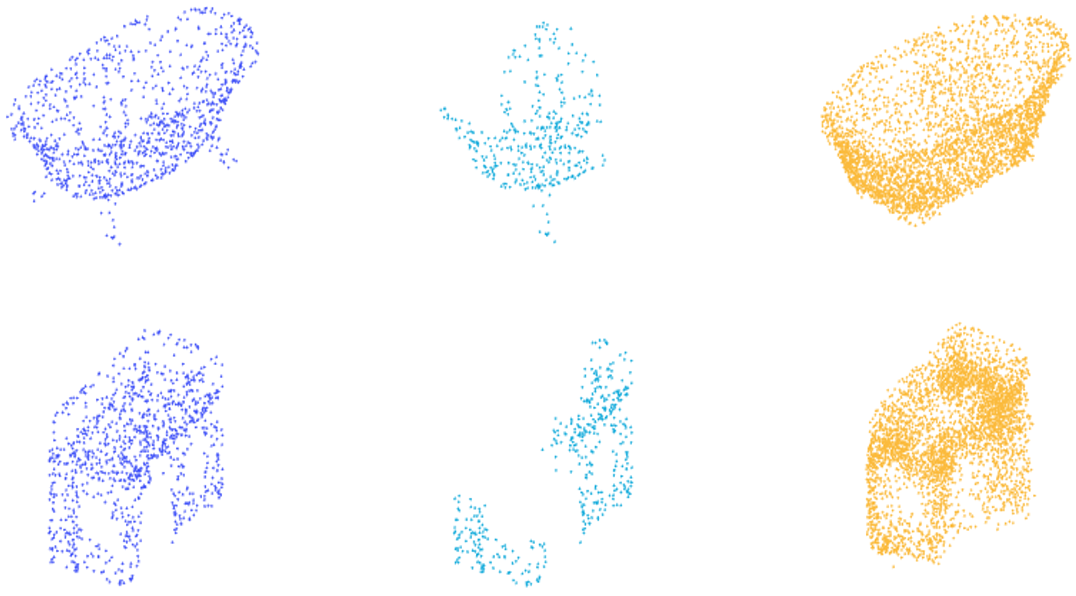}
        \end{minipage}}
    \caption{Visualization of reconstructed point cloud samples with random masking in the target domain of PointDA-10.} 
    \label{fig:reconstruction}
\end{figure}

It is noteworthy that for the challenging yet realistically significant Simulation-to-Reality scenarios (\ie \textbf{M} $\rightarrow$ \textbf{S*},  \textbf{S} $\rightarrow$ \textbf{S*}, \textbf{M11} $\rightarrow$ \textbf{SO*11}, and \textbf{S9} $\rightarrow$ \textbf{SO*9}), our proposed RPD acquires a remarkable enhancement over w/o Adapt by 21.1\%, 18.6\%, 15.37\%, and 20.08\% respectively. Visualization of confusion matrices in terms of class-wise classification accuracy achieved by the w/o Adapt and our RPD on four Simulation-to-Reality UDA tasks are shown in Fig. \ref{Fig.confusion_matrix}.

\subsection{Visualization}
\begin{figure}[ht]
    \centering
    \subfigure[PointDAN]{
    \begin{minipage}[b]{0.1\textwidth}
        \label{Fig.PointDAN}
            \includegraphics[width=\textwidth]{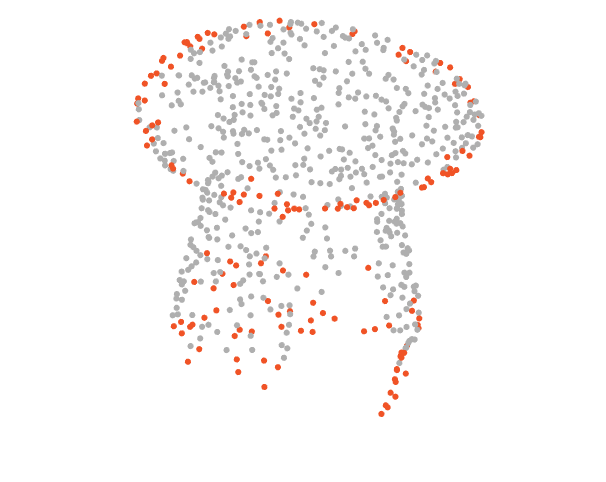}
            \includegraphics[width=\textwidth]{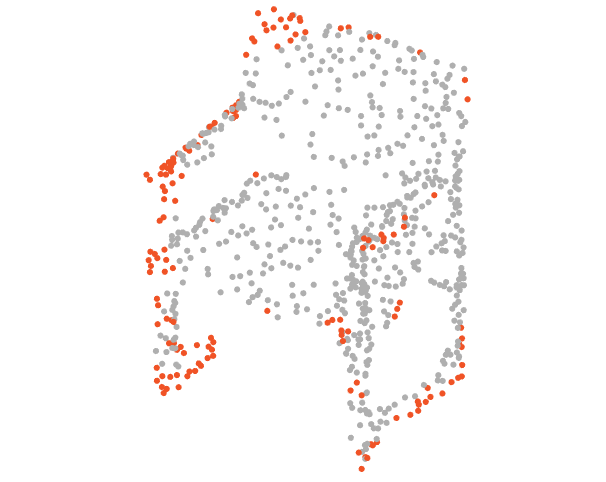}
            \includegraphics[width=\textwidth]{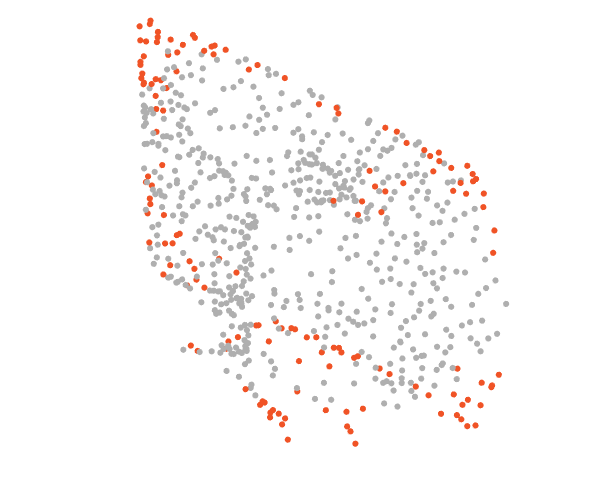}
    \end{minipage}
    }
    \subfigure[DefRec]{
    \begin{minipage}[b]{0.1\textwidth}
        \label{Fig.DefREC}
            \includegraphics[width=\textwidth]{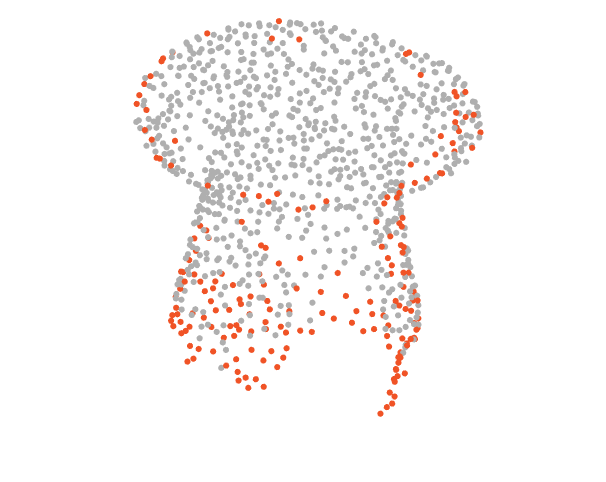}
            \includegraphics[width=\textwidth]{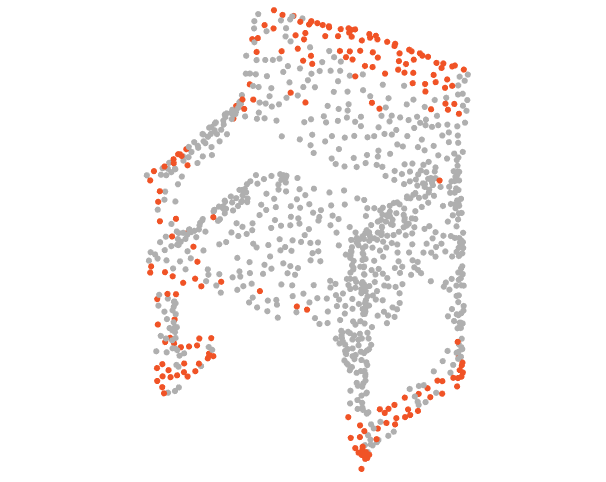}
            \includegraphics[width=\textwidth]{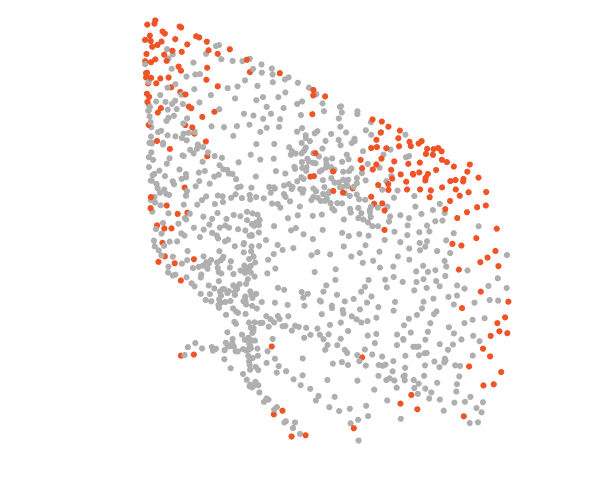}
    \end{minipage}
    }
    \subfigure[SD]{
    \begin{minipage}[b]{0.1\textwidth}
        \label{Fig.SD}
            \includegraphics[width=\textwidth]{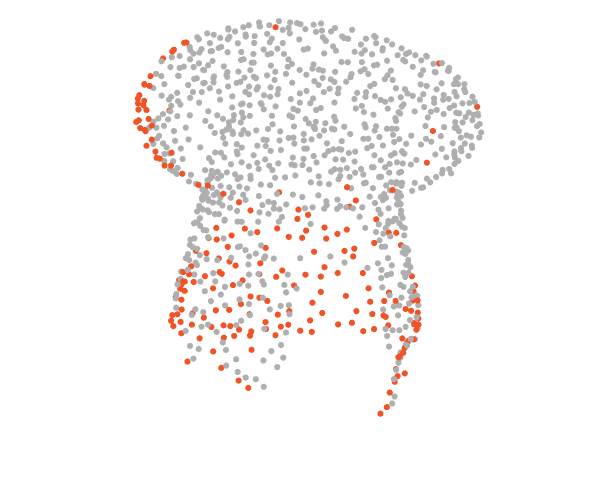}
            \includegraphics[width=\textwidth]{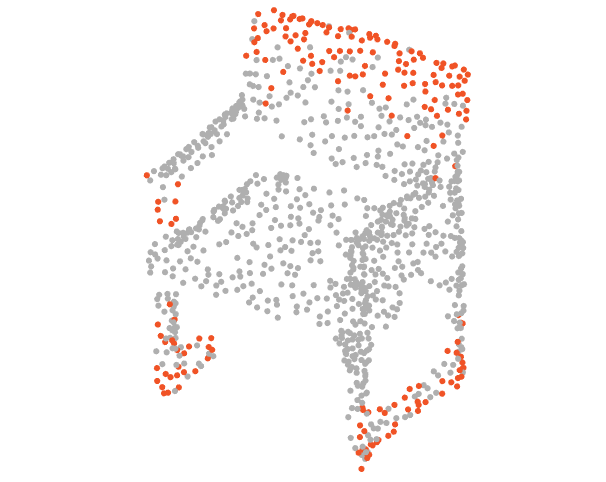}
            \includegraphics[width=\textwidth]{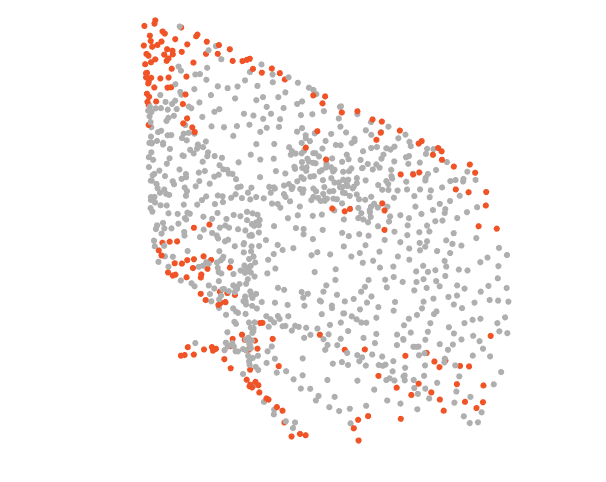}
    \end{minipage}
    }
    \subfigure[RPD]{
    \begin{minipage}[b]{0.1\textwidth}
        \label{Fig.RPD}
            \includegraphics[width=\textwidth]{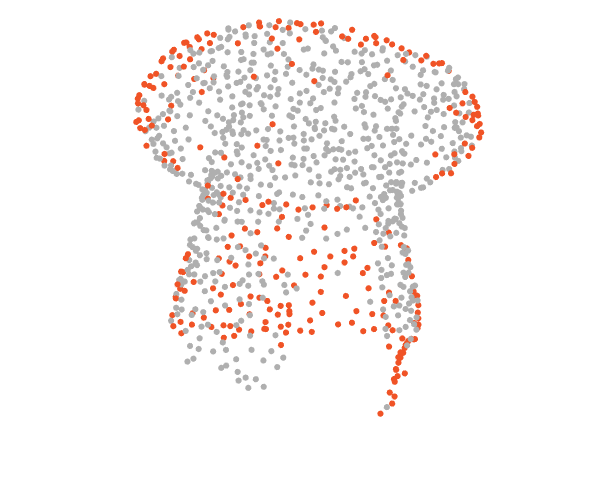}
            \includegraphics[width=\textwidth]{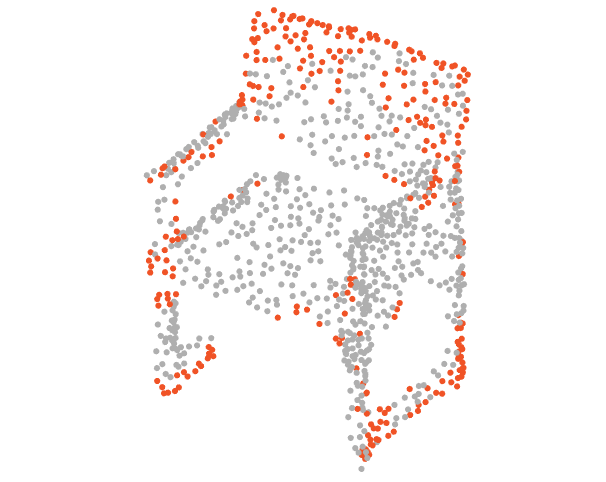}
            \includegraphics[width=\textwidth]{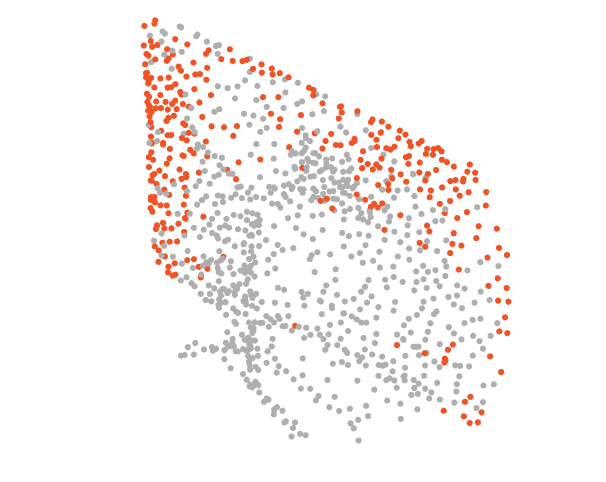}
    \end{minipage}
    }
    \caption{Saliency map visualization of various comparison methods under the setting of M$\rightarrow$S* on PointDA-10.} 
    \label{fig:saliency_map}
\end{figure}

We visualize the input point clouds, random masking, and the reconstructed 3D coordinates in Fig. \ref{fig:reconstruction}. We believe that reconstruct masked point cloud with masked 2D image tokens can create a challenging self-supervised learning task that encourage the network to learn 3D geometric information. We use the saliency map analysis method referenced in \cite{pc_saliency} to visualize the features of various comparison methods under the setting of M$\rightarrow$S* on PointDA-10 \cite{PointDAN}. As shown in Figure \ref{fig:saliency_map}, the proposed RPD method utilizes prior knowledge to model the relationships between local patches using pre-trained Vision Transformers (ViT). The proposed RPD focuses on the global structure and effectively captures the local relationships within point cloud data, enabling our method to extract robust and invariant 3D semantic representations. In contrast, most other comparison methods use DGCNN as the backbone for feature extraction, which tends to focus only on local high-frequency areas, such as edges, thereby affecting generalization.

\section{Discussion}
\subsection{Scalability}
As shown in Table \ref{tab:vit-compares}, we experimented with three different scales of pre-trained ViT models: ViT-S, ViT-B and ViT-L. However, we observed that using the larger-scale pre-trained model did not lead to performance improvements.

\begin{table*}[ht]
\centering
\caption{Results on the effect of using different pre-trained 2D Transformer-based models on PointDA-10.}
\renewcommand{\arraystretch}{1.15}
\resizebox{0.7\linewidth}{!}{
\begin{tabular}
{l|l|ccccccc}
\hline
Domain & Methods & M$\rightarrow$S & M$\rightarrow$S* & S$\rightarrow$M & S$\rightarrow$S* & S*$\rightarrow$M & S*$\rightarrow$S & Avg \\
\hline
\multirow{3}{*}{Source}
& RPD-S & 98.7 & 98.8 & 94.5 & 94.3 & 76.5 & 78.0 & 90.3 \\
& RPD-B & 98.7 & 98.9 & 94.0 & 94.7 & 77.4 & 79.3 & 90.7 \\
& RPD-L & \textbf{99.0} & \textbf{99.3} & \textbf{95.4} & \textbf{95.7} & \textbf{78.1} & \textbf{80.0} & \textbf{91.3} \\
\hline
\multirow{3}{*}{Target}
& RPD-S & 80.4 & 63.5 & 81.7 & 58.4 & 74.5 & 73.5 & 72.0 \\
& RPD-B & \textbf{81.9} & \textbf{64.4} & \textbf{82.8} & \textbf{59.0} & \textbf{77.1} & \textbf{76.4} & \textbf{73.6} \\
& RPD-L & 80.9 & 61.7 & 80.4 & 58.9 & 75.0 & 75.9 & 72.1 \\
\hline
\end{tabular}}
\label{tab:vit-compares}
\end{table*}

We attribute this to several factors: First, our relatively small point cloud dataset is prone to overfitting, which is exacerbated by larger ViT-L models. Second, we use only 27 patches compared to the 196 used in ViT-B. This smaller number of patches means that larger ViT-L models are not needed for effective patch relationship modeling. Finally, using larger ViT-L models would require more patches, potentially leading to insufficient information in each patch and reducing the effectiveness of local relationship modeling.

\subsection{Limitation}
While our RPD approach demonstrates considerable promise, there are certain limitations worth for investigation:

\noindent \textbf{Computational Complexity:}
As shown in Fig. \ref{tab:param_anals}, the use of pre-trained ViT model leads to increased computational requirements, especially when dealing with very large datasets or high-resolution point clouds. Future work could focus on optimizing the model for efficiency or exploring more lightweight architectures that retain the robustness of pre-trained ViT model.

\noindent \textbf{Effects of Openset Data:}
The pre-trained ViT model is based on 2D data, which might influence our results. If the 2D pre-training data does not include any sample of semantic categories of the PointDA-10 \cite{PointDAN} and Sim-to-Real \cite{MetaSets} point cloud datasets, our method's performance could be adversely affected. The absence of relevant categories in the pre-training dataset can result in suboptimal feature extraction and limited generalization.

\noindent \textbf{Effects of Pre-training Approaches:}
Different pre-training approaches for ViT \cite{vit}, such as DINO \cite{DINO} and MAE \cite{MAE}, can have varying impacts on the performance of our method. We have observed that MAE pre-trained ViT tends to have an advantage in our experiments. The pre-training method affects the quality of learned representations and the model’s effectiveness in downstream tasks.

\noindent \textbf{Extension to Other 3D Data Types:}
While our current focus is on point clouds, extending the proposed RPD to other types of 3D data, such as voxel grids or mesh data, could enhance its applicability. Adapting and optimizing the approach for these data types is an interesting area for future work.

\begin{table}[ht]
\centering
\caption{Comparative analysis of training costs of different methods}
\renewcommand{\arraystretch}{1.15}
\resizebox{0.8\linewidth}{!}{
\begin{tabular}{lcc}
\hline
Methods & Parameters (M) & FLOPs(G) \\ 
\hline
DefRec \cite{DefRec} & 2.08 & 2.77 \\
PointDAN \cite{PointDAN} & 2.84 & 0.94\\
SD \cite{SD} & 3.47 & 0.92 \\
GAI \cite{GAI} & 22.68 & 3.58 \\
GAST \cite{GAST} & 23.60 & 2.17 \\
RPD (ours) & 62.27 & 23.29 \\ 
\hline
\end{tabular}}
\label{tab:param_anals}
\end{table}

\section{Conclusion}
This paper proposes a novel scheme for unsupervised domain adaptation on object point cloud classification, aiming to alleviate domain shift by distilling relational priors from pre-trained 2D transformers.
We illustrate how the relational priors learned by a proficient 2D Transformer model can be transferred to the 3D domain, thereby enhancing the generalization of 3D features. Our methodology involves adopting a standard teacher-student distillation framework, where the parameter-frozen pre-trained Transformer module is shared between the 2D teacher model and the 3D student model. Additionally, we employ an online knowledge distillation strategy to further semantically regularize the 3D student model. Moreover, to empower the model's capacity to capture 3D geometric information, we introduce a novel self-supervised task involving the reconstruction of masked point cloud patches using corresponding masked multi-view image features. Experiments conducted on two public benchmarks validate the efficacy of our approach, demonstrating new state-of-the-art performance.

\bibliographystyle{IEEEtran}
\bibliography{IEEEabrv, Main.bib}

\vfill

\end{document}